\documentclass[a4paper,fleqn,final]{cas-sc}
\usepackage[utf8]{inputenc}

\usepackage[numbers]{natbib}

\def\tsc#1{\csdef{#1}{\textsc{\lowercase{#1}}\xspace}}
\tsc{WGM}
\tsc{QE}
\tsc{EP}
\tsc{PMS}
\tsc{BEC}
\tsc{DE}

\usepackage[utf8]{inputenc}
\usepackage[T1]{fontenc}
\usepackage{textcomp}
\usepackage{amssymb}
\usepackage{longtable}
\usepackage{array}
\usepackage{colortbl}
\usepackage{amsmath}
\usepackage{graphicx}
\usepackage[inline]{enumitem}
\usepackage{ragged2e}
\usepackage{placeins}
\usepackage{threeparttable}
\usepackage{hyperref}
\hypersetup{
colorlinks   = true,
urlcolor     = purple,
linkcolor    = purple,
citecolor   = purple}
\makeatletter
    \g@addto@macro{\UrlBreaks}{\do\/\do\-\do\_} 
\makeatother
\usepackage[ocgcolorlinks]{ocgx2}
\usepackage{multirow}
\usepackage{float}
\usepackage{placeins}
\usepackage{afterpage}
\usepackage{appendix}
\usepackage{enumitem}
\usepackage{subcaption} 
\usepackage[skins,breakable]{tcolorbox}
\colorlet{shadecolor}{yellow}
\usepackage{mdwmath}
\usepackage{mdwtab}
\usepackage{eqparbox}
\usepackage{booktabs}
\usepackage{multirow}
\usepackage{numprint}
\definecolor{lightblue}{rgb}{0.92, 0.95, 1}
\definecolor{lightred}{rgb}{1, 0.9, 0.9}
\definecolor{deepblue}{rgb}{0, 0.4470, 0.7410}
\definecolor{deepyellow}{rgb}{0.9290, 0.6940, 0.1250}
\definecolor{deepgreen}{rgb}{0,0.5,0}
\definecolor{mygreen}{rgb}{0.01, 0.5, 0.01}
\definecolor{myred}{rgb}{0.8, 0.01, 0.01}
\usepackage{longtable}
\usepackage{url}
\usepackage{soul}
\usepackage{xcolor}

\usepackage{algorithm}
\usepackage{algpseudocode}
\usepackage{amsmath}
\usepackage{graphicx}

\usepackage{setspace}

\newdefinition{rmk}{Remark}

\newproof{intuition}{intuition}

\newproof{pot}{Proof of Theorem \ref{thm}}

\definecolor{deepgreen}{rgb}{0.0, 0.5, 0.0}

\begin{document}
\let\WriteBookmarks\relax
\renewcommand{\floatpagefraction}{0.7}
\renewcommand{\textfraction}{0.1}
\renewcommand{\topfraction}{0.9}
\renewcommand{\bottomfraction}{0.9}
\setcounter{topnumber}{4}
\setcounter{bottomnumber}{2}
\setcounter{totalnumber}{6}

\shorttitle{}

\shortauthors{Nasif, Jahin, \& Mridha (2026)}

\title[mode = title]{Reinforcement-Guided Hyper-Heuristic Hyperparameter Optimization for Fair and Explainable Spiking Neural Network-Based Financial Fraud Detection}








\author[1]{Sadman Mohammad Nasif}[orcid=0009-0000-5079-541X]
\ead{nasif1907120@stud.kuet.ac.bd}
\fnmark[1]
\credit{Investigation, Writing -- original draft, Software}

\author[2]{Md Abrar Jahin}[orcid=0000-0002-1623-3859]
\ead{jahin@usc.edu}
\fnmark[1]
\credit{Conceptualization, Project Administration, Formal analysis, Investigation, Methodology, Software, Writing -- original draft, Visualization}

\author[3]{M. F. Mridha}[orcid=0000-0001-5738-1631]
\ead{firoz.mridha@aiub.edu}
\cormark[1]
\credit{Supervision, Validation, Writing -- review \& editing}

\cortext[cor1]{Corresponding author(s)}
\fntext[1]{These authors contributed equally to this work.}

\affiliation[1]{organization={Department of Computer Science \& Engineering, Khulna University of Engineering \& Technology (KUET)},
    city={Khulna},
    postcode={9203},
    country={Bangladesh}}

\affiliation[2]{organization={Thomas Lord Department of Computer Science, Viterbi School of Engineering, University of Southern California},
    city={Los Angeles},
    state={CA},
    postcode={90089},
    country={USA}}

\affiliation[3]{organization={Department of Computer Science, American International University-Bangladesh (AIUB)},
    city={Dhaka},
    postcode={1229},
    country={Bangladesh}}

\begin{abstract}
The growing adoption of home banking systems has heightened the risk of cyberfraud, necessitating fraud detection mechanisms that are not only accurate but also fair and explainable. While AI models have shown promise in this domain, they face key limitations, including computational inefficiency, the interpretability challenges of spiking neural networks (SNNs), and the complexity and convergence instability of hyper-heuristic reinforcement learning (RL)-based hyperparameter optimization. To address these issues, we propose a novel framework that integrates a Cortical Spiking Network with Population Coding (CSNPC) and a Reinforcement-Guided Hyper-Heuristic Optimizer for Spiking Systems (RHOSS). The CSNPC, a biologically inspired SNN, employs population coding for robust classification, while RHOSS uses Q-learning to dynamically select low-level heuristics for hyperparameter optimization under fairness and recall constraints. Embedded within the Modular Supervisory Framework for Spiking Network Training and Interpretation (MoSSTI), the system incorporates explainable AI (XAI) techniques, specifically, saliency-based attribution and spike activity profiling, to increase transparency. Evaluated on the Bank Account Fraud (BAF) dataset suite, our model achieves a 90.8\% recall at a strict 5\% false positive rate (FPR), outperforming state-of-the-art spiking and non-spiking models while maintaining over 98\% predictive equality across key demographic attributes. Although RHOSS introduces an offline search cost, this is amortized over deployment, where inference dominates. The architectural sparsity of CSNPC ensures lower energy per transaction compared to dense ANNs, as supported by neuromorphic benchmarks. The explainability module further confirms that saliency attributions align with spiking dynamics, validating interpretability. These results demonstrate the potential of combining population-coded SNNs with reinforcement-guided hyper-heuristics for fair, transparent, and high-performance fraud detection in real-world financial applications.
\end{abstract}


\maketitle

\section{Introduction}
\label{sec:introduction}
Home banking systems have become increasingly popular over the past few years, driven by technological advancements and the growing demand for convenient and accessible financial services. But with this increased reliance on digital platforms comes the additional risk of cyberfraud, which has led to significant economic losses within the sector. Due to the effectiveness in detecting such frauds, artificial intelligence (AI) algorithms have become a common security measure for home banking systems~\citep{bdcc7020093, RYMANTUBB2018130}. Still, the AI algorithms come with their own set of drawbacks, primarily their energy inefficiencies due to the computational intensity involved in processing vast amounts of data and floating-point operations~\citep{snnsurvey, rozenberg_computing_2012}. Spiking Neural Networks (SNNs) emerge as a potential solution for this challenge, offering an energy-efficient solution while maintaining the high performance required for fraud detection tasks~\citep{maass_networks_1997}.

SNNs mimic the brain as a complex network of neurons that interact through electrochemical signals and an interconnected population of cells that process information~\citep{hao_biologically_2020, eshraghian, wang_hippocampal_2025, chen2025}. Due to their event-driven nature and sparse architecture, when combined with neuromorphic hardware, they can be implemented in a far more energy-efficient system compared to traditional Artificial Neural Networks (ANNs). Recent research has increasingly focused on refining SNN learning mechanisms, aiming to replicate the activation of brain neurons, where an entire area of the brain is activated with synaptic impulses of spiking neurons~\citep{Davies2018,tang2025}. This coordinated activity could be achieved through population coding, where groups of neurons collectively represent and process information, offering improved robustness, efficiency, and accuracy compared to the simplistic approach of assigning one neuron per target class in Machine Learning (ML) problems~\citep{neurondynamics, maass_networks_1997}. This could lead to increased efficiency and enable SNNs to address practical challenges, such as handling highly imbalanced dataset~\citep{qiao2023}. With this breakthrough, SNNs could open new possibilities for applications across various domains, including fraud detection, although challenges remain in optimizing these networks for real-world scenarios. 

One of the major challenges of an SNN model is the training process, during which the model must optimize its hyperparameters. This is where Hyper Heuristic Reinforcement Learning (HHRL) comes into play. HHRL is a method that combines hyperheuristics with reinforcement learning (RL) to optimize decision-making in complex environments. Hyperheuristics are specialized algorithms that generate or select heuristics, problem-specific strategies, or parameters used to solve computational tasks. When integrated with reinforcement learning, it can dynamically select and refine these heuristics based on continuous feedback from the environment~\citep{choong_automatic_2018}. This combination enables the system to evolve and improve over time, making it highly effective for solving hyperparameter optimization problems where traditional heuristics-based optimizers often fall short~\citep{zhang_deep_2022}. The primary advantage of HHRL is its ability to handle uncertainty and complex problem spaces by learning to adapt and select the most suitable heuristics, ultimately leading to improved performance and flexibility in the model.

To ensure these complex models remain understandable, transparent, and trustworthy in real-world applications, explainability becomes a critical component of the overall framework. Explainable AI (XAI) refers to techniques in artificial intelligence that make the decision-making and predictions of machine learning models more accessible or understandable to humans~\citep{Hsu04052023}. As current models become increasingly complex, the nature of these systems continues to elude the grasp of users and clients. XAI provides techniques such as visualization, rule-based explanation, or human-understandable features to provide interpretations to users or clients. The use of XAI, along with visualization techniques and explanations, also has the added benefit of facilitating model debugging and ensuring fairness and ethical decision-making~\citep{hoffman2019, holzinger2022}. 

In this study, we develop a model that incorporates HHRL with \textit{\ul{C}ortical \ul{S}piking \ul{N}etwork with \ul{P}opulation \ul{C}oding} (CSNPC), a biologically inspired convolutional spiking neural network (CSNN). Our model improves financial fraud detection within a real-world benchmark problem using the Bank Account Fraud (BAF) dataset suite~\citep{jesus2022}. The model is embedded within the \textit{\ul{Mo}dular \ul{S}upervisory Framework for \ul{S}piking Network \ul{T}raining and \ul{I}nterpretation} (MoSSTI), which enables our model to be more explainable through saliency-based attribution and spike activity profiling. Additionally, our model accounts for variations in real-world client demographics and ensures fairness for any client, regardless of their demographic differences~\citep{Barocas2018FairnessAM}. Another improvement of our model is observed in its false positive rate (FPR) threshold of below 5\%, which enables the model to reduce the impact on general clients, making it suitable for deployment in the real world. Again, population coding improves robustness and fairness by distributing representation across neurons, reducing sensitivity to demographic bias. RHOSS formalizes hyperparameter optimization as a Markov Decision Process, enabling dynamic adaptation under fairness constraints, an approach not explored in prior work.

The main contributions of this study are:
\begin{enumerate}
  \item We propose CSNPC, a cortical spiking neural network with population coding that improves robustness, fairness, and recall for imbalanced fraud detection. Unlike prior SNNs, CSNPC leverages structured spiking outputs to improve class-level separability while adhering to strict 5\% FPR constraints.
  \item We introduce \textit{\ul{R}einforcement-Guided \ul{H}yper-Heuristic \ul{O}ptimizer for \ul{S}piking \ul{S}ystems} (RHOSS), a novel Q-learning-based hyper-heuristic optimization framework that dynamically selects low-level heuristic (LLH) operations to search the spiking hyperparameter space. RHOSS enables fairness-aware, reward-driven exploration, outperforming static and Bayesian methods in convergence stability and constraint satisfaction.
  \item We integrate CSNPC and RHOSS into the MoSSTI framework, which provides a unified training and interpretability pipeline. By combining saliency-based attribution with spike activity profiling, MoSSTI enables dual-path explainability, revealing strong correspondence between spiking behavior and feature relevance.
  \item We conduct a rigorous evaluation on the BAF dataset suite, covering six fairness-sensitive variants and three demographic attributes (age, income, employment). Our method consistently demonstrates generalization, ethical alignment, and better performance compared to spiking and classical baselines under real-world constraints.
\end{enumerate}

The article is structured in six sections. Section \hyperref[sec:litreview]{2} presents a general overview of SNNs and past works on similar models. Section \hyperref[sec:proposed_approach]{3} describes our model and how it functions. Section \hyperref[sec:experimantal_setup]{4} explains the utilized resources, datasets, and evaluation parameters for our experiments. Section \hyperref[sec:results_and_discussions]{5} describes the experimental results and their comparison with other works. Section \hyperref[sec:conclude]{6} concludes the paper and identifies areas where opportunities for improvements in the future exist.

\section{Literature Review}
\label{sec:litreview}
Cyberfraud has become a major concern for financial institutions due to the rise of home banking systems. Fraud detection has been a point of interest for researchers for quite some time, especially in the field of Artificial Intelligence. The recent BAF dataset developed by Jesus et al. has been used to improve fraud detection when it comes to highly imbalanced data, as is common in the real-world client demography~\citep{jesus2022}. Additionally, the BAF dataset sets a bias toward groups of people, making it a benchmark in assessing the performance and fairness of the models. This study will utilize this dataset for its experiments and evaluations. 

Initial assessments of the BAF dataset were performed by multiple studies using decision tree and gradient boosting algorithms~\citep{ke2017lightgbm}. Pombal et al. used 6 different ML algorithms: XGBoost, LightGBM, Logistic Regression, Decision Tree, Random Forest, and Feed Forward Neural Network trained with the Adam optimizer and achieved a result varying between 25\% and 75\% of recall at 5\% of FPR, and fairness between 30\% and 75\% of predictive equality, depending on the dataset variant~\citep{understanding_unfairness}. Later, Ding et al. introduced Fair Gradient-Boosting Machine (FairGBM)~\citep{cruz2023fairgbm}, which uses Lagrangian-constrained optimization to ensure fairness. However, this model was only tested on datasets such as the Account Opening Fraud (AOF) and other American Community Survey (ACS)-based datasets~\citep{ding2022retiringadultnewdatasets}. Uwoma analyzed multiple algorithms and assessed their strengths, limitations, and fairness~\citep{Uwaoma}. Despite fine-tuning the models, when a strict 5\% FPR threshold is enforced, LightGBM and Random Forest showcased 47\% and 48\% recall values, which means the models still require significant improvements to perform fairly in real-world environments. Luzio et al.~\citep{Luzio2024} deployed three models: Catboost~\citep{CatBoost}, LightGBM, and Multi-Layered Perceptron (MLP)~\citep{Popescu2009} and improved the performance. Although their work did not measure fairness, their calibration strategies increased performance, reaching recall values of 52\% with Catboost, 54\% with LightGBM, and 49\% with the MLP. Yousefimehr et al. ~\citep{yousefimehr2025} presents a hybrid fraud detection framework combining One-Class Support Vector Machine (OCSVM) with the Synthetic Minority Oversampling Technique (SMOTE) and random undersampling, evaluated using LightGBM and LSTM models. It achieves 87\% F1 score and 96\% AUC scores on the European credit card dataset. But it should be noted that the European credit card dataset is outdated compared to the BAF dataset, which is ideal for the evaluation of these models. 

In response to this challenge, SNNs are emerging as a promising solution. Recent studies have explored the potential of SNNs from computational neuroscience and neuromorphic perspectives, indicating their ability to address the complexities of real-world problems. For example, Hao et al. \citep{hao_biologically_2020} proposed a method that integrates symmetric spike-timing dependent plasticity, inspired by neuroscience. Zhang et al.~\citep{Zhang2019} employed a new learning method for spiking neurons, a membrane potential-driven approach, where the postsynaptic membrane potential, rather than the postsynaptic spike times, serves as the key signal for synaptic modifications. 

Recent studies employed SNNs with the BAF dataset in an effort to improve fraud detection results and the fairness of the models. Perdig\~{a}o et al.~\citep{dylan_spiking_csnnpc} used SNNs, leveraging population coding for improved performance and fairness. The authors optimized a 1D-Convolutional SNN with Bayesian optimization on the BAF dataset. Their approach achieved 47.08\% recall at a 5\% FPR, outperforming traditional models like LightGBM while ensuring above 90\% fairness across sensitive attributes. Similarly, Ribeiro et al.~\citep{dylan_bptt} introduce SpikeConv, a CSNN using LIF neurons for fraud detection. Utilizing Optuna for hyperparameter optimization, their model SpikeConv achieved 56.88\% Recall and 94.79\% Accuracy while keeping FPR below 5\%. 

To conclude, financial fraud detection using machine learning remains an ever-challenging field for ML researchers. The BAF dataset has become a widely adopted benchmark for improving fraud detection performance. However, only a limited number of studies have focused on addressing fairness in their algorithms. There is still a need for further research to improve fairness while ensuring the optimization of recall at a 5\% FPR, which is a standard business constraint for models used in high-stakes financial applications. Although previous studies employ SNNs with some success, their choices of hyperparameter optimizers do not fully exploit the potential of advanced optimization techniques, potentially limiting the performance and fairness improvements. Despite the relative success of these studies, they also fail to integrate Explainable AI (XAI) techniques, which are essential for enhancing model transparency, interpretability, and trustworthiness. Utilizing explainable AI can provide additional benefits, such as easier debugging and insights into the model’s performance and where it is succeeding or failing.

\begin{figure}[pos=h]
    \centering
    \includegraphics[width=1\linewidth]{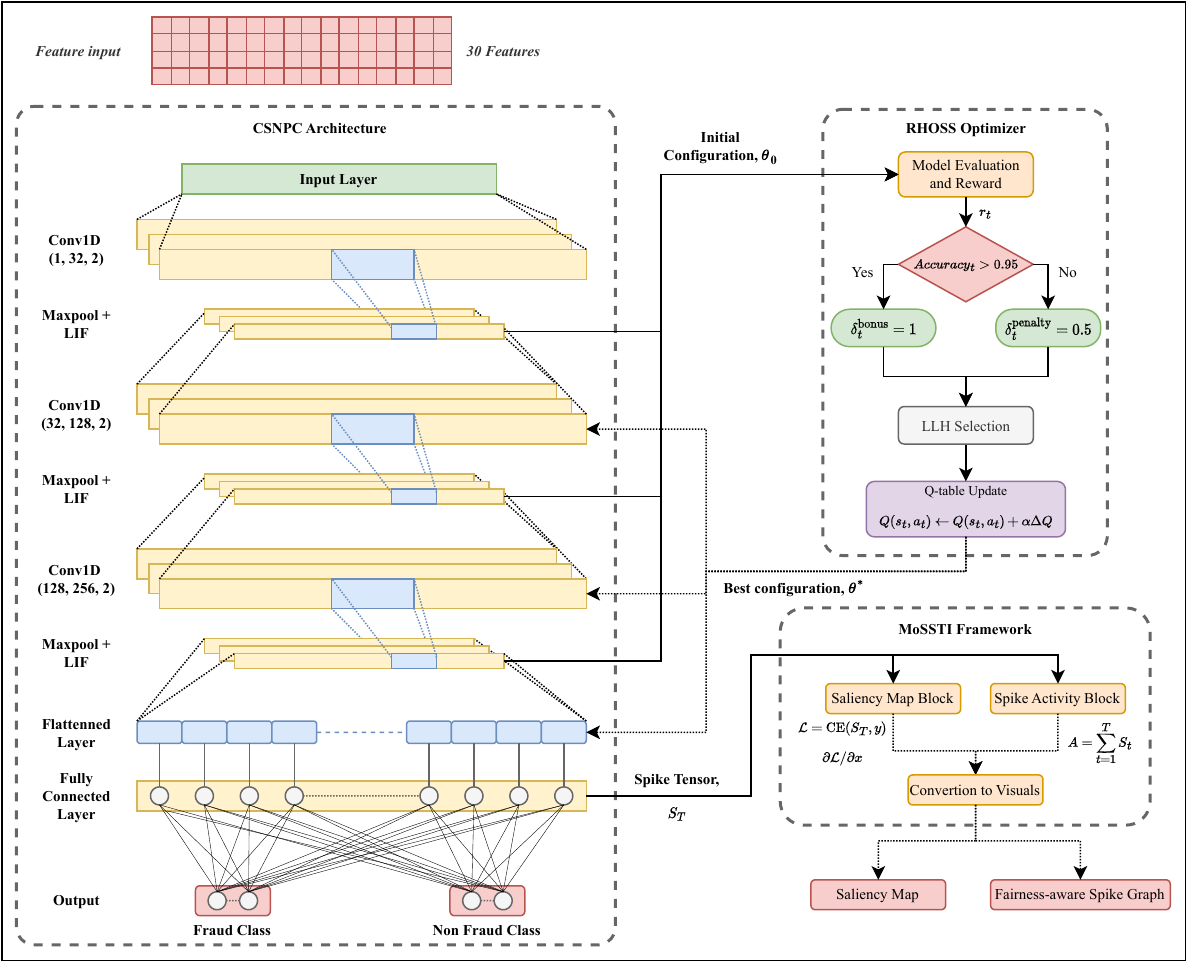}
    \caption{Overview of the proposed Reinforcement-Guided Hyper-Heuristic Optimization framework. The MoSSTI integrates the CSNPC model with RHOSS, a Q-learning-based optimizer. The workflow illustrates how model configurations are evaluated via model performance and fairness metrics. Based on evaluation outcomes, rewards are computed, and the RL agent adaptively selects LLHs to guide the hyperparameter search process. The Q-table is updated using a temporal-difference learning rule to progressively improve decision policies. The framework incorporates explainability through saliency-based and spike-activity analysis modules to ensure fairness-aware, interpretable fraud detection.}
    \label{fig:framework}
\end{figure}

\section{Proposed Approach}
\label{sec:proposed_approach}
Our proposed approach employs a two-stage process to optimize and evaluate the CSNPC, a biologically inspired 1D CSNN. The architecture consists of three convolutional layers with 32, 128, and 256 filters (kernel size 2), each followed by maximum pooling and LIF neurons to capture spatiotemporal dynamics across $T$ simulation steps. The output is flattened and projected to a population of $P$ spiking neurons, with $P/2$ neurons assigned per class, enabling prediction via population spike summation. The model is embedded within the MoSSTI training framework, which enables it to be explained through saliency-based attribution and spike activity profiling. To optimize model performance under fairness and recall constraints, we introduce RHOSS, a Q-learning-based hyperheuristic optimizer that adaptively selects low-level heuristics to explore the hyperparameter space. This unified framework enables the development of explainable and high-performing spiking models tailored for real-world fraud detection. The complete integration of CSNPC and RHOSS within the MoSSTI framework, along with its optimization workflow and explainability modules, is illustrated in Figure~\ref{fig:framework}.

\subsection{Model Architecture}
The core of the proposed model architecture is the CSNPC, a class-based CSNN designed to leverage both spatial feature extraction and temporal spike-based dynamics. This architecture is deployed within a modular supervisory framework referred to as the MoSSTI, which governs the complete learning pipeline, including training, inference, and post hoc interpretability. The model processes input samples formatted as feature vectors over a discrete sequence of simulation steps, thereby capturing the dynamic evolution of neuronal spiking activity. Each input instance is represented as a three-dimensional tensor, with dimensions corresponding to batch size, input channel, and feature dimensionality. This temporal decomposition enables the spiking model to encode discriminative temporal patterns, reflecting the inherent timing-based representations fundamental to spiking neural networks.

The CSNPC model is made up of a hierarchy of convolutional and spiking layers. Initially, the input is passed through a one-dimensional convolutional layer with 32 filters and a kernel size of 2, followed by a max-pooling operation that halves the temporal resolution. This is followed by a LIF neuron layer, which models the subthreshold membrane dynamics and spike generation. The LIF neuron integrates its membrane potential according to the update equation $u_t = \beta u_{t-1} + I_t - S_{t-1} V_{\text{th}}$, where $u_t$ is the membrane potential at time step $t$, $\beta$ is a learnable decay constant, $I_t$ is the input current, $S_{t-1}$ is the spike generated at the previous time step, and $V_{\text{th}}$ is a learnable threshold. When $u_t \geq V_{\text{th}}$, a spike is emitted. This structure is repeated two more times with increasing convolutional filters, 128 and 256, respectively, each followed by max pooling and a corresponding LIF layer, resulting in a deep feature hierarchy.

The final spiking feature map is flattened and passed through a fully connected layer that projects the signal to a spiking population of neurons. These output neurons are also LIF units with trainable parameters and are configured to emit spike responses that represent class probabilities through population coding. In population coding, each class is represented by a group of neurons within the output population, and the class prediction is made by summing the spikes of each group and selecting the class with the highest total spike count over all time steps: 
\begin{equation}
\hat{y}_c = \arg\max_{c \in C} \sum_{t=1}^{T} S_t^{(c)}
\end{equation}
where $S_t^{(c)}$ denotes the spike activity of neurons assigned to class $c$ at time $t$.

The entire network is unrolled over $T$ timesteps, and at each step, the data flows through the static convolutional layers and the dynamic LIF layers. At the same time, the spike and membrane states are updated and recorded. The detailed sequence of operations performed in the forward pass is described in Algorithm~\ref{alg:CSNPC}.

\begin{algorithm}[H]
\caption{Forward Pass of the CSNPC Model}
\label{alg:CSNPC}
\begin{algorithmic}[1]
\State \textbf{Initialize} membrane potentials $u_1$, $u_2$, $u_3$, $u_4$ for each LIF layer
\For{each time step $t \in \{1, \dots, T\}$}
    \State Apply 1D convolution: $x_1 = \text{Conv1d}(x)$
    \State Apply max pooling: $x_2 = \text{MaxPool}(x_1)$
    \State Compute spike and membrane update: $(S_1, u_1) = \text{LIF}(x_2, u_1)$
    \State Apply convolution and pooling on $S_1$: $(S_2, u_2)$ via next LIF layer
    \State Repeat convolution and pooling to obtain $(S_3, u_3)$
    \State Flatten $S_3$ and apply linear projection: $z = \text{FC}(S_3)$
    \State Compute output spike: $(S_4, u_4) = \text{LIF}(z, u_4)$
    \State Record current, spike, and membrane outputs at this time step
\EndFor
\State \Return Stacked current, spike, and membrane tensors across time
\end{algorithmic}
\end{algorithm}

This architecture is embedded within the \textit{MoSSTI} framework, which automatically configures the training parameters, initializes the optimizer, and handles class imbalance via weighted spike count cross-entropy loss. The model prediction is made by decoding the most active class using spike summation. This integrated setup enables seamless support for downstream explainable AI modules such as saliency map extraction and spike activity analysis.

\subsection{XAI Integration}
To facilitate transparency in the decision-making process of the spiking neural model, we incorporated two complementary XAI techniques within the \textit{MoSSTI} framework: saliency-based attribution and spike activity profiling. These methods were applied post hoc to trained models to extract feature relevance and temporal spiking dynamics for individual input samples.

Saliency maps are generated by computing the gradient of the network's output probability for the input features. Let $x$ denote the input vector and $\mathcal{L}$ be the loss function associated with the predicted output for a target class. The saliency map is computed as the absolute gradient of the loss concerning the input, $\left|\partial \mathcal{L} / \partial x\right|$, which identifies which input dimensions most strongly influenced the prediction. To achieve this, the model first activates evaluation mode and receives the input sample as a tensor with gradients enabled. The forward pass is performed through the simulation steps $T$ to produce a final spike count tensor $S_T$. The predicted loss is calculated by cross-entropy between $S_T$ and the true class label. A backward pass is then executed to propagate the gradients, and the resulting input gradient is extracted and returned as the saliency map. This entire gradient-based explanation process is formalized in Algorithm~\ref{alg:saliency}.

\begin{algorithm}[H]
\caption{Saliency-Based Explanation}
\label{alg:saliency}
\begin{algorithmic}[1]
\State \textbf{Set} model to evaluation mode
\State \textbf{Convert} input $x$ to a differentiable tensor and move to device
\State \textbf{Convert} label $y$ to tensor on device
\State \textbf{Perform} forward pass through network to get spike record $S$
\State \textbf{Extract} final spike output $S_T$ at last timestep
\State \textbf{Compute} loss $\mathcal{L} = \text{CE}(S_T, y)$
\State \textbf{Backpropagate} to compute gradients: $\partial \mathcal{L} / \partial x$
\State \textbf{Return} absolute gradient as saliency map: $\left|\partial \mathcal{L} / \partial x\right|$
\end{algorithmic}
\end{algorithm}

In addition to saliency, we analyze the spike activity at the population level to assess the encoding strength of each feature. After converting the input sample into a spiking format and running it through the trained network, the spike tensor is summed over the full simulation window. The resulting vector represents the total spike count per output neuron, which can be mapped back to feature dimensions by population coding. This provides a biologically grounded, explainable signal that reveals which features sustained higher spiking responses over time. The spike activity computation is summarized in Algorithm~\ref{alg:spikeactivity}.

\begin{algorithm}[H]
\caption{Spike Activity by Feature}
\label{alg:spikeactivity}
\begin{algorithmic}[1]
\State \textbf{Set} model to evaluation mode
\State \textbf{Convert} input $x$ to tensor and move to device
\State \textbf{Perform} forward pass to obtain spike record $S$
\State \textbf{Sum} spikes over time: $A = \sum_{t=1}^{T} S_t$
\State \textbf{Return} spike activity vector $A$
\end{algorithmic}
\end{algorithm}

To provide a unified XAI interface, a combined method was implemented that jointly returns both the saliency gradient and the spike activity. This dual perspective allows model predictions to be explained both through classical backpropagation and through temporal neuron firing dynamics. These XAI diagnostics were evaluated using the best-performing configuration (\textit{P200-S20}) on the benchmark datasets under fairness and business constraints, allowing the insights from individual samples to be contextualized with broader behavioral metrics across sensitive subgroups.

\begin{table}[!t]
\centering
\caption{Hyperparameter Search Space for SNN Optimization}
\label{tab:searchspace}
\begin{tabular}{lcc}
\toprule
\textbf{Hyperparameter} & \textbf{Range} & \textbf{Sampling Distribution} \\
\midrule
Membrane decay ($\beta_L$) & $ 0.1-0.95 $ & Log-uniform \\
Spike slope ($\sigma$) & $ 10-50 $ & Log-uniform \\
Adaptive threshold ($\theta_L$) & $0.1-1.0$ & Log-uniform \\
Weight decay ($\omega$) &  $0.95-1.0 $ & Log-uniform \\
Adam betas ($\beta_k$) & $0.97-0.99$ & Uniform \\
Learning rate ($\lambda$) & $10^{-6}-10^{-3}$ & Log-uniform \\
\bottomrule
\end{tabular}
\end{table}

\begin{algorithm}[H]
\caption{RHOSS for SNN Hyperparameter Optimization}
\label{alg:hyperheuristic}
\begin{algorithmic}[1]
\State Initialize Q-table: $Q(s, a) \leftarrow 0$ for all $s \in \{0,\dots,4\}, a \in \{0,\dots,9\}$
\State Set $\epsilon \leftarrow 1.0$, $\alpha \leftarrow 0.1$, $\gamma \leftarrow 0.9$
\State Generate initial configuration $\theta_0$ using LLH0
\State Evaluate $\theta_0$ to obtain reward $r_0$
\State Set best configuration $\theta^* \leftarrow \theta_0$, $r^* \leftarrow r_0$
\For{$t = 1$ to $T$}
    \State $s_t \leftarrow \lfloor \frac{t}{T} \cdot 5 \rfloor$
    \State Select action $a_t$ using $\epsilon$-greedy policy on $Q(s_t, \cdot)$
    \State Apply LLH $a_t$ to $\theta^*$ to generate candidate $\theta_t$
    \State Train SNN using $\theta_t$ and compute reward $r_t$
    \If{$r_t > r^*$}
        \State Update best: $\theta^* \leftarrow \theta_t$, $r^* \leftarrow r_t$
    \EndIf
    \State $s_{t+1} \leftarrow \lfloor \frac{t+1}{T} \cdot 5 \rfloor$
    \State Update Q-table:
    \[
    Q(s_t, a_t) \leftarrow Q(s_t, a_t) + \alpha \left( r_t + \gamma \max_{a'} Q(s_{t+1}, a') - Q(s_t, a_t) \right)
    \]
    \State Decay $\epsilon \leftarrow \max(0.05, \epsilon \cdot 0.99)$
\EndFor
\State \Return Best configuration $\theta^*$ and associated metrics
\end{algorithmic}
\end{algorithm}

\subsection{Reinforcement-Guided Hyper-Heuristic Optimizer for Spiking Systems} 
\label{sec:hh_rl}
To effectively optimize the hyperparameters of the SNN, we propose RHOSS, a Q-learning-based hyperheuristic optimizer. This approach utilizes a Q-learning agent that adaptively selects from a portfolio of low-level heuristics, which are designed to either sample new configurations or perturb existing ones. By alternating between exploration and exploitation strategies, the agent efficiently navigates the high-dimensional hyperparameter space. The optimization process is formalized as a Markov Decision Process (MDP), where each episode corresponds to a single trial of hyperparameter optimization. The discrete state space is partitioned into five phases, representing different portions of the search horizon. The action space comprises ten LLHs, indexed from 0 to 9. LLH0 performs stochastic initialization by sampling new configurations, while LLH1 to LLH9 apply targeted perturbations to specific hyperparameters relevant to SNN dynamics.

At each time step $t$, the agent observes the current state $s_t$, selects an action $a_t$ according to an $\epsilon$-greedy policy, and applies the chosen LLH to the current best configuration $\theta^*$. This produces a candidate configuration $\theta_t$, which is evaluated by training an SNN and computing a scalar reward $r_t$:
\begin{equation}\label{eq:2}
r_t = \text{Recall}_t - \text{FPR}_t + \delta_t^{\text{bonus}} - \delta_t^{\text{penalty}}
\end{equation}
where $\delta_t^{\text{bonus}} = 1$ if $\text{Accuracy}_t > 0.95$, and $\delta_t^{\text{penalty}} = 0.5$ if no positive instances are correctly identified in the test set despite their presence. The Q-values are updated using the standard temporal difference learning rule:
\begin{equation}
Q(s_t, a_t) \leftarrow Q(s_t, a_t) + \alpha \left( r_t + \gamma \max_{a'} Q(s_{t+1}, a') - Q(s_t, a_t) \right)
\end{equation}
where $\alpha = 0.1$ is the learning rate, $\gamma = 0.9$ is the discount factor, and the exploration rate $\epsilon$ decays linearly from 1.0 to 0.05 throughout training. The hyperparameter search space is summarized in Table~\ref{tab:searchspace}. A mixture of log-uniform and uniform sampling strategies is employed to accommodate the varying scales and sensitivity of different parameters. The full procedure is outlined in Algorithm~\ref{alg:hyperheuristic}, which details the sequential decision-making steps, LLH application, reward computation, and Q-table update.

Unlike static or Bayesian optimizers, RHOSS leverages RL to adaptively select low-level heuristics based on reward signals that encode both performance and fairness. This dynamic search strategy theoretically improves constraint satisfaction compared to fixed policies. This reinforcement-driven optimization strategy facilitates the automated discovery of robust and generalizable hyperparameter configurations. By integrating domain-informed low-level heuristics within a dynamic learning framework, RHOSS shows improved adaptability and performance across real-world fraud detection tasks. Furthermore, it effectively aligns with fairness-aware objectives and recall-oriented performance targets.

\begin{table}[!ht]
\centering
\caption{Best hyperparameters resulting from the RHOSS-based optimization process}
\label{tab:3}
\begin{tabular}{lccccccc}
\toprule
\textbf{Hyperparameters} & \textbf{Symbol} & \multicolumn{3}{c}{\textbf{20 Steps}} & \multicolumn{3}{c}{\textbf{50 Steps}} \\
\cmidrule(lr){3-5} \cmidrule(lr){6-8}
 & & \textbf{P = 2} & \textbf{P = 20} & \textbf{P = 200} & \textbf{P = 2} & \textbf{P = 20} & \textbf{P = 200} \\
\midrule
\multirow{4}{*}{Decays} 
 & $\beta_1$ & 0.1185 & 0.9012 & \textbf{0.1185} & 0.1525 & 0.2324 & 0.2324 \\
 & $\beta_2$ & 0.1990 & 0.1215 & \textbf{0.1990} & 0.9500 & 0.8502 & 0.8502 \\
 & $\beta_3$ & 0.1537 & 0.9500 & \textbf{0.1537} & 0.3339 & 0.5196 & 0.5196 \\
 & $\beta_4$ & 0.1830 & 0.2877 & \textbf{0.1830} & 0.3434 & 0.3849 & 0.3849 \\
\midrule
Slope & $\sigma$ & 15 & 15 & \textbf{12} & 27 & 48 & 48 \\
\midrule
\multirow{4}{*}{Thresholds}
 & $\theta_1$ & 0.1477 & 0.3182 & \textbf{0.1477} & 0.6854 & 0.2791 & 0.2791 \\
 & $\theta_2$ & 0.3604 & 0.1126 & \textbf{0.3604} & 0.9499 & 0.1259 & 0.1259 \\
 & $\theta_3$ & 0.8633 & 0.1900 & \textbf{0.8633} & 0.6862 & 0.2879 & 0.2879 \\
 & $\theta_4$ & 0.4966 & 0.8096 & \textbf{0.4966} & 0.1305 & 0.2156 & 0.2156 \\
\midrule
Adam Weight & $\omega$ & 0.9782 & 0.9618 & \textbf{0.9782} & 0.9548 & 0.9570 & 0.9570 \\
\multirow{2}{*}{Adam Betas} 
 & $\beta_1$ & 0.993 & 0.971 & \textbf{0.993} & 0.989 & 0.972 & 0.972 \\
 & $\beta_2$ & 0.987 & 0.990 & \textbf{0.987} & 0.997 & 0.991 & 0.991 \\
Adam Learning Rate & $\lambda$ & 7.00e$^{-5}$ & 2.30e$^{-5}$ & \textbf{7.00e$^{-5}$} & 2.10e$^{-5}$ & 1.00e$^{-6}$ & 1.00e$^{-6}$ \\
\bottomrule
\end{tabular}
\end{table}

\begin{table*}[!ht]
\caption{Best-performing configurations for each variant based on classification performance metrics. This table reports the selected best model per variant by jointly optimizing for low FPR and FNR, and high TPR (Recall), TNR, and accuracy. \textbf{Bold} indicates the best performance for each hyperparameter configuration across all the metrics. \ul{Underline} indicates the best overall performance with a business restriction of 5\% FPR for each variant}
\label{tab:benchmark}
\resizebox{\textwidth}{!}{%
\begin{tabular}{@{}lcccccccc@{}}
\toprule
\multicolumn{1}{l}{\textbf{Configuration}} &
  \textbf{FPR ($\downarrow$)} &
  \textbf{TPR ($\uparrow$)} &
  \textbf{TNR ($\uparrow$)} &
  \textbf{FNR ($\downarrow$)} &
  \textbf{Accuracy ($\uparrow$)} &
  \textbf{Training Time (min) ($\downarrow$)} &
  \textbf{Inference Time (min) ($\downarrow$)} &
  \textbf{Data Variant} \\ \midrule
\multirow{6}{*}{P200-S50} & 0.156 & 0.487 & 0.844 & 0.513 & 0.986 & 1815.704 & 31.409 & Base          \\
                          & 0.024 & 0.613 & 0.756 & 0.387 & 0.986 & 1792.296 & 30.221 & Variant I     \\
                          & \textbf{0.137} & 0.457 & \textbf{0.863} & 0.543 & 0.986 & 1793.883 & 29.247 & Variant II    \\
                          & 0.441 & \textbf{0.900}   & 0.559 & \textbf{0.100}   & \textbf{\textbf{0.989}} & 1723.405 & 35.008 & Variant   III \\
                          & 0.152 & 0.454 & 0.848 & 0.546 & \textbf{\textbf{0.989}} & 1713.76  & 36.482 & Variant IV    \\
                          & 0.192 & 0.535 & 0.808 & 0.465 & \textbf{\textbf{0.989}} & 1735.778 & 37.195 & Variant V     \\ \midrule
\multirow{6}{*}{P2-S20}   & \textbf{0.045} & 0.319 & \textbf{0.955} & 0.681 & 0.946 & 710.465  & 13.542 & Base          \\
                          & 0.612 & \textbf{0.957} & 0.388 & \textbf{0.043} & 0.986 & 701.198  & 13.033 & Variant I     \\
                          & 0.111 & 0.586 & 0.889 & 0.414 & 0.986 & 704.532  & 13.364 & Variant II    \\
                          & 0.131 & 0.481 & 0.869 & 0.519 & \textbf{0.989} & 674.569  & 14.995 & Variant   III \\
                          & 0.000  & 0.000  & 0.000  & 1.000  & \textbf{0.989} & 666.095  & 15.542 & Variant IV    \\
                          & 0.048 & 0.339 & 0.952 & 0.661 & 0.946 & 677.799  & 14.772 & Variant V     \\ \midrule
\multirow{6}{*}{\textcolor{myred}{P200-S20}} & \ul{\textbf{0.014}} & \ul{0.476} & \ul{0.857} & \ul{0.524} & \ul{0.986}  & 710.136  & 13.035 & Base          \\
                          & \ul{0.047} & \ul{\textbf{0.908}} & \ul{0.526} & \ul{\textbf{0.092}} & \ul{0.986} & 711.352  & 12.975 & Variant I     \\
                          
                          & \ul{0.080}  & \ul{0.489} & \ul{0.920}  & \ul{0.511} & \ul{0.986}  & 712.427  & 12.903 & Variant II    \\
                          
                          & \ul{0.025} & \ul{0.601} & \ul{0.753} & \ul{0.399} & \ul{\textbf{0.989}} & 680.97   & 15.27  & Variant   III \\
                          & 0.029 & 0.063 & \textbf{0.971} & 0.937 & 0.961& 678.696  & 15.997 & Variant IV    \\
                          
                          & \ul{0.045} & \ul{0.853} & \ul{0.552} & \ul{0.147} & \ul{\textbf{0.989}} & 681      & 15.43  & Variant V     \\ \midrule
\multirow{6}{*}{P20-S20}  & \textbf{0.097} & 0.386 & 0.903 & 0.614 & 0.986& 712.865  & 13.502 & Base          \\
                          & 0.034 & \textbf{0.874} & 0.662 & \textbf{0.126} & 0.986 & 709.369  & 13.335 & Variant I     \\
                          & 0.128 & 0.588 & 0.872 & 0.412 & 0.986 & 707.24   & 13.412 & Variant II    \\
                          & 0.115 & 0.447 & 0.885 & 0.553 & \textbf{0.989} & 674.98   & 15.515 & Variant   III \\
                          & 0.000     & 0.000     & \textbf{1.000}     & 1.000     & \textbf{0.989} & 677.263  & 15.41  & Variant IV    \\
                          & 0.039 & 0.826 & 0.615 & 0.174 & \textbf{0.989} & 679.851  & 15.566 & Variant V     \\ \midrule
\multirow{6}{*}{P20-S50}  & 0.053 & 0.165 & 0.947 & 0.835 & 0.986 & 1739.942 & 30.139 & Base          \\
                          & 0.107 & 0.385 & 0.893 & 0.615 & 0.986 & 1739.028 & 29.468 & Variant I     \\
                          & 0.006 & 0.025 & 0.994 & 0.975 & 0.981 & 1750.228 & 29.823 & Variant II    \\
                          & 0.131 & 0.401 & 0.869 & 0.599 & \textbf{0.989} & 1684.705 & 35.859 & Variant   III \\
                          & \textbf{0.002} & 0.005 & \textbf{0.998} & 0.995 & 0.988 & 1672.15  & 34.492 & Variant IV    \\
                          & 0.028 & \textbf{0.781} & 0.722 & \textbf{0.219} & \textbf{0.989} & 1676.582 & 35.025 & Variant V     \\ \midrule
\multirow{6}{*}{P2-S50}   & 0.090  & 0.359 & 0.910  & 0.641 & 0.986 & 1757.293 & 29.781 & Base          \\
                          & 0.000     & 0.000     & \textbf{1.000}     & 1.000     & 0.986 & 1752.085 & 29.807 & Variant I     \\
                          & 0.103 & 0.385 & 0.897 & 0.615 & 0.986 & 1761.833 & 30.444 & Variant II    \\
                          & \textbf{0.009} & 0.027 & 0.991 & 0.973 & 0.981 & 1679.506 & 35.813 & Variant   III \\
                          & 0.112 & 0.378 & 0.888 & 0.622 & \textbf{0.989} & 1656.299 & 36.18  & Variant IV    \\
                          & 0.054 & \textbf{0.910}  & 0.459 & \textbf{0.090}  & \textbf{0.989} & 1708.544 & 36.534 & Variant V     \\ \bottomrule
\end{tabular}%
}

\smallskip
\footnotesize \textit{Note:} Training time includes the complete RHOSS optimization process (100 trials × average 6.8 min per trial) plus final model training. Inference time represents processing the full 100K test set.
\end{table*}

\begin{table}[!ht]
\centering
\caption{Predictive equalities, model parameters, and corresponding reward for the best trials across six hyperparameter configurations}
\label{tab:4}
\begin{tabular}{lccccc}
\toprule
\textbf{Configuration} 
& {\textbf{$PE_{Age}$} ($\uparrow$)} 
& {\textbf{$PE_{Income}$} ($\uparrow$)} 
& {\textbf{$PE_{Employment}$} ($\uparrow$)} 
& {\textbf{\#Params} ($\downarrow$)} 
& {\textbf{Reward}} \\
\midrule
P200-S20 & 0.99037 & 0.98071 & 0.95927 & 228016 & 0.30728 \\
P2-S20   & 0.98508 & 0.98907 & 0.97532 & \textbf{75754}  & 0.29091 \\
P20-S20  & 0.98855 & 0.97075 & 0.95365 & 89596  & 0.29115 \\
\textbf{P2-S50}   & \textbf{0.99174} & 0.98870 & \textbf{0.98059} & \textbf{75754}  & 0.19986 \\
P20-S50  & 0.96960 & \textbf{0.99407} & 0.90116 & 89596  & 1.07599 \\
P200-S50 & 0.00000 & 0.00000 & 0.00000 & 228016 & 0.31018 \\
\bottomrule
\end{tabular}
\end{table}

\subsection{Synergistic Integration and Computational Complexity Analysis}
The integration of CSNPC with RHOSS creates synergistic advantages that exceed the sum of individual components.

\textbf{Theoretical Justification.}
Population coding in CSNPC creates a richer hyperparameter landscape with smoother gradients, enabling more effective exploration by RHOSS. Specifically, the use of $P$ neurons per class leads to a $P$-dimensional representation space, which reduces the likelihood of local minima common in single-neuron-per-class SNNs. RHOSS's Q-learning mechanism further exploits this structure through adaptive heuristic selection, allowing it to navigate the search space more efficiently.

\textbf{Why This Combination Works.}
First, \emph{the search space structure} becomes more tractable: population coding transforms discrete classification into a continuous spike-count optimization problem, which Q-learning handles more effectively than  Bayesian methods operating on discrete spaces. Second, \emph{fairness-aware optimization} is achieved through RHOSS’s reward function (Eq.~\ref{eq:2}), which directly incorporates fairness metrics. This enables joint optimization of performance and equity, an objective that static optimizers cannot handle. Finally, \emph{temporal dynamics} inherent in SNNs introduce optimization challenges due to their multi-step temporal evolution ($T$ steps). RHOSS’s state-based approach, comprising five distinct phases, naturally aligns with temporal credit assignment mechanisms in spiking networks, enhancing stability and convergence.

\textbf{Computational Complexity.}
The overall training complexity is $\mathcal{O}(T \times N \times E \times H)$, where $T$ denotes optimization trials (100), $N$ is the number of samples (900K), $E$ the number of epochs (20), and $H$ the hyperparameter dimensionality (14). Inference complexity is $\mathcal{O}(S \times N \times L)$, with $S$ being the simulation steps (20) and $L$ the number of layers (4). The computational investment during training, approximately $\mathcal{O}(10^9)$ operations, is amortized across millions of inference operations ($\mathcal{O}(10^6)$), yielding a favorable time–performance trade-off suitable for deployment scenarios.

\section{Experimental Setup}
\label{sec:experimantal_setup} 
The experimental evaluation was conducted on a server equipped with NVIDIA T4 Tensor Core GPUs (2,560 CUDA cores and 320 Tensor Cores) with reproducible random seeds for statistical validation. A multitude of specialized software libraries were employed to support various stages of model development and data processing. Aequitas version 1.0.0~\citep{aequitas} was utilized for group-wise disparity analysis, while CTGAN~\citep{ctgan} was used in the original construction of the BAF dataset. No explicit resampling or class balancing techniques were applied during training; instead, class imbalance was handled via a weighted loss function. Visualizations were created using matplotlib~\citep{Matplotlib} and seaborn~\citep{seaborn}. Model training was performed using PyTorch~\citep{pytorch}, scikit-learn~\citep{scikit-learn}, and SciPy ~\citep{SciPy}. Finally, spiking neural networks were assessed using the different BAF dataset variants using the snnTorch3 Python library~\citep{snntorch3}.

\subsection{Benchmark Dataset and Preprocessing}
\label{subsec:benchmark dataset}

\subsubsection{Dataset Description}
The \textit{BAF Dataset Suite}~\citep{jesus2022} serves as a fairness-oriented benchmark, encompassing six variations from real-world online bank account application processes. To preserve client anonymity, synthetic data was generated using a Generative Adversarial Network (GAN). The dataset includes 30 features spanning eight months of data, totaling 1,000,000 instances, with 90\% allocated for training (covering six months) and 10\% reserved for testing (spanning two months). Each variant introduces distinct types of bias and modeling challenges, making the suite suitable for evaluating multiple machine learning approaches. Each dataset variant includes three sensitive attributes: age, income, and employment status, with group-wise fraud prevalence detailed in Fig. 3. For this study, the sensitive groups are defined using three criteria: age, income, and employment status. Individuals aged 50 years or older are classified as "older," while those younger than 50 are considered "younger". Similarly, individuals with an income value greater than or equal to 0.5 are categorized as "rich," while those with a value below 0.5 are categorized as "poor."  Individuals with an employment status value greater than or equal to 3 are labeled "unstable," and those below 3 as "stable." These groupings facilitate the evaluation of fairness across demographic segments using the FPR ratios, also referred to as predictive equality \citep{algorithmic_decision_making}. 

Variant I creates an imbalance in group sizes, where the minority group is decreased from 20\% to 10\% of the overall dataset. Both groups retain an equal fraud prevalence rate of 1.1\% despite this. Consequently, models trained on this variant are not exposed to prevalence differences but must be resilient to the underrepresentation of the minority group. Variant II is more imbalanced in terms of prevalence compared to Variant I and the original data. In this case, while the group sizes are balanced at 50\%, the fraud rate is 0.4\% for the majority group and 1.9\% for the minority group. This setup serves as a baseline for evaluating model fairness when there are significant prevalence imbalances. Variant III focuses on separability imbalance. It maintains equal group sizes (50\%) and identical fraud prevalence (1.1\%) across groups. However, the majority group is favored by two additional synthetic features that improve class separability, which creates a favorable condition for the majority group in model training. Variant IV replicates a prevalence disparity similar to Variant II, with fraud rates of 0.3\% for the majority and 1.7\% for the minority group. When temporal aspects are considered, the two groups are equalized to an adjusted fraud prevalence of 1.5\%. This reflects real-world scenarios where time-based biases in data collection influence model behavior, demonstrating the necessity to consider temporal changes in fairness measurements. Variant V also addresses temporal bias, but from the angle of separability. Both group sizes and fraud prevalence rates are equal at 50\% and 1.1\%, respectively. However, for the initial six months, there is a benefit of higher separability for the majority group with additional features. In the latter half, this benefit deteriorates, and models must adapt to shifting patterns of bias between training and test periods. 

\subsubsection{Dataset Split, Preprocessing and Augmentation}
We strictly adhere to the BAF Dataset Suite's predefined 90-10 split, with 900,000 samples allocated for training (months 1-6) and 100,000 for testing (months 7-8). This temporal split preserves the realistic scenario in which models must generalize to future, unseen fraud patterns. In accordance with the BAF suite protocol, we conducted minimal preprocessing, which involved several key steps. First, we applied min-max scaling to normalize all 30 numerical features to a range of [0, 1]. We also opted against using synthetic oversampling techniques, such as SMOTE, to preserve the realistic class imbalance, as the fraud rate was only 1.1\% in most variants. We utilized the original 30 features without creating any additional features to ensure a fair comparison with baseline models.

\paragraph{Rationale for No Augmentation.}
The only data augmentation mentioned in our pipeline (Section 4) refers to CTGAN being used to generate the BAF dataset itself \cite{jesus2022}, not our training process. We intentionally avoided class-balancing techniques because: (i) Previous SNN studies \cite{dylan_spiking_csnnpc,dylan_bptt} also did not use oversampling; (ii) Population coding handles imbalance through neuron redundancy; (iii) Preserves realistic deployment conditions. Our comparisons in Table~\ref{tab:comparison} are fair because all methods operate on the same imbalanced data without augmentation.

\subsection{Model Evaluation}
\label{subsec:model_evaluation}
To evaluate the performance of our model in real-world applications, we utilize multiple performance metrics and compare them with those of previous methods and experiments. True Positive Rate (TPR) measures the model's effectiveness in correctly identifying positive instances. Similarly, the True Negative Rate (TNR) measures the correct identification of negative instances, indicating that the system reduces false alarms and incorrect identifications. On the other hand, the False Positive Rate (FPR) indicates the rate at which the model incorrectly flags a negative instance as positive, resulting in a higher rate of false alarms. Similarly, the False Negative Rate (FNR) is the rate at which the model incorrectly flags a positive instance as negative, causing the positive instance to slip under the radar. Accuracy is the measure of the proportion of instances where the system identifies both positives and negatives correctly, and is usually a metric through which we can judge a model's overall performance.

In addition to these metrics, we assess the fairness of the models using the Predictive Equality metric \citep{algorithmic_decision_making, understanding_unfairness}, which is computed as the ratio of FPRs between two groups ($g_1$, $g_2$) for each sensitive attribute $\text{attr} \in \{\text{customer age}, \text{income}, \text{employment status}\}$.
\begin{equation}
\text{PE}_{\text{attr}} =
\begin{cases}
\frac{\text{FPR}_{g_1}}{\text{FPR}_{g_2}}, & \text{if } \text{FPR}_{g_2} \geq \text{FPR}_{g_1} \\
\frac{\text{FPR}_{g_2}}{\text{FPR}_{g_1}}, & \text{otherwise}
\end{cases}
\end{equation}

This formulation enables us to assess disparities in fraud prediction across sensitive demographic groups using the Aequitas Python library \citep{aequitas}. A higher value of $\text{PE}_{\text{attr}}$ indicates greater fairness for the attribute. This is illustrated in Figure~\ref{fig:spike_activity}, which shows the model's performance (recall) and fairness across various population sizes. Finally, we compute the trade-off $T_{\text{attr}}$ between model performance and fairness for each sensitive attribute as:
\begin{equation}
T_{\text{attr}} = \alpha \cdot \text{TPR@5FPR} + (1 - \alpha) \cdot \text{PE}_{\text{attr}}
\end{equation}
where $\alpha \in [0, 1]$ controls the relative importance of performance versus fairness. A value of $\alpha=0.5$ represents equal weighting of both criteria.



\begin{table}[!t]
\centering
\caption{Comparison with the state-of-the-art models on BAF dataset under a 5\% FPR constraint}
\label{tab:comparison}
\begin{tabular}{lcccccc}
\toprule
\textbf{Configuration} & \textbf{Model} & \textbf{\#Params} & \textbf{Algorithm} & \textbf{FPR} & \textbf{Recall} & \textbf{Accuracy (\%)} \\
\midrule
DSNN & 1 hidden~\citep{dylan_bptt}  & 170002   & BPTT         & 0.087      & 0.290      & 90.31 \\
DSNN & 2 hidden~\citep{dylan_bptt}   & 267502   & BPTT         & 0.046      & 0.210      & 94.34 \\
SpikeConv & M3~\citep{dylan_bptt}     & 38544    & TBPTT         & 0.045      & 0.280      & 94.50 \\
SpikeConv & M5~\citep{dylan_bptt}     & 38544    & BPTT         & 0.050      & 0.570      & 94.79 \\
GBDT & LightGBM~\citep{dylan_bptt}   & --       & GBDT         & 0.037      & 0.450      & --    \\
\midrule
P200-S50 & CSNN~\citep{dylan_spiking_csnnpc} & --    & --           & 0.043     & 0.471     & --    \\
\midrule
P200-S20 & CSNPC (Ours)              & 711352   & RHOSS        & 0.047      & \textbf{0.908} & \textbf{98.60} \\
\bottomrule
\end{tabular}
\end{table}



\section{Results and Discussions}
\label{sec:results_and_discussions}
This section presents a consolidated evaluation of the proposed CSNPC, optimized via RHOSS, and interpreted using MoSSTI. The evaluation spans predictive performance, fairness trade-offs, and explainability, benchmarked across six spatiotemporal encoding configurations defined by time steps \( S \in \{20, 50\} \) and population sizes \( P \in \{2, 20, 200\} \).

\subsection{Performance and Fairness Trade-offs}
In the first phase of model training, RHOSS was used to optimize our model's hyperparameters with \( P = 200, s = 20\), showing the most optimal results. The reward function combines recall and FPR with fairness-aware bonuses, aligning optimization with business and ethical constraints. This design ensures convergence toward configurations that satisfy strict FPR limits while maintaining predictive equality. Table \ref{tab:3} illustrates how varying population sizes (P) and optimization steps (20 or 50) impact the hyperparameter values in the RHOSS-based optimization process for training the model. Comparing the evaluation parameters across six configurations in Table \ref{tab:benchmark}, only the configuration \( P=200, S=20 \) adheres to a 5\% FPR across all dataset variants. This threshold ensures that the model does not incorrectly flag too many instances as positive, making the model compatible with the banking and finance sector's standards.  The configuration \( P=200, S=20 \) performs best with dataset variant II, where it achieves a recall of 0.908 and an FNR of 0.092.  During training, our model directly accounted for model fairness (Predictive Equality) and adjusted the training parameters accordingly. Although RHOSS incurs significant offline compute cost, this is justified by its ability to enforce fairness constraints systematically, which is a critical requirement in financial applications. Table \ref{tab:4} illustrates how the predictive equalities (age, income, and employment status)  were considered when determining reward values for each configuration of our model. The configuration \(P=2, S=50\) shows better predictive equality overall with 99.17\% for $PE_{Age}$ and 98.4\% for $PE_{Employment}$,  while using the fewest number of parameters among the configurations, making it more energy-efficient.

\begin{figure}[pos=h]
\centering
\begin{subfigure}{0.30\linewidth}
    \centering
    \includegraphics[width=\linewidth]{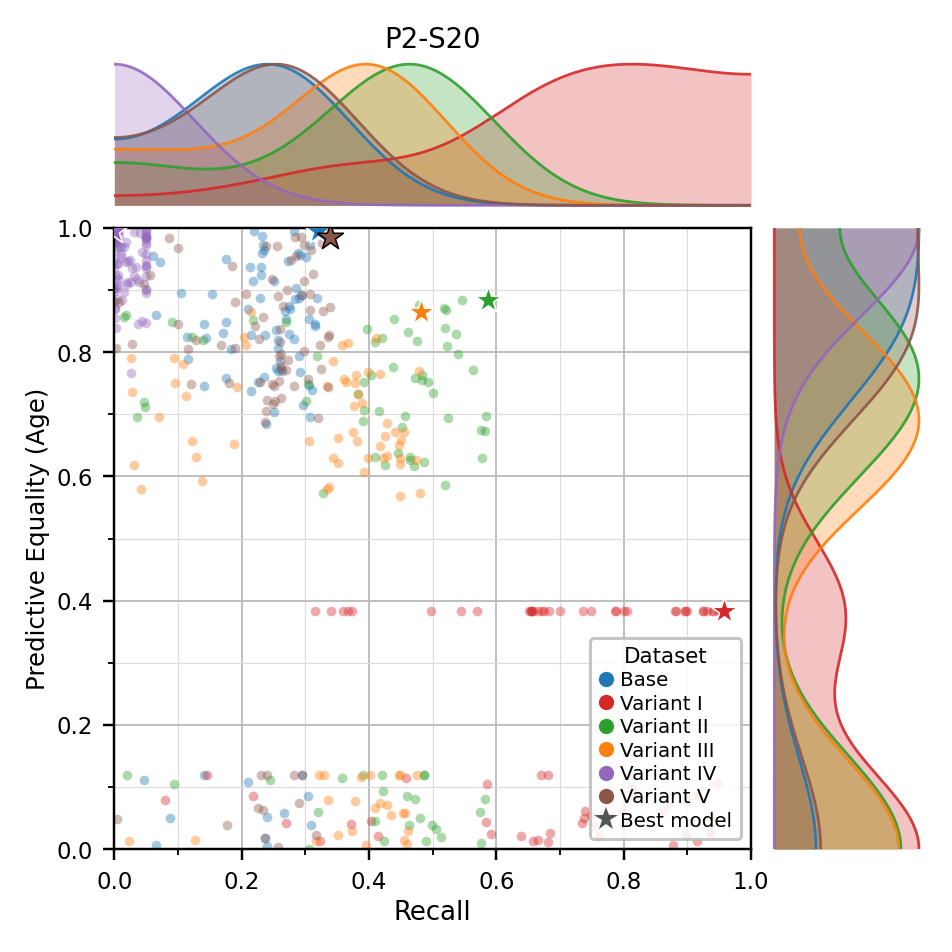}
    \caption{P2-S20 -- Age}
\end{subfigure}
\hfill
\begin{subfigure}{0.30\linewidth}
    \centering
    \includegraphics[width=\linewidth]{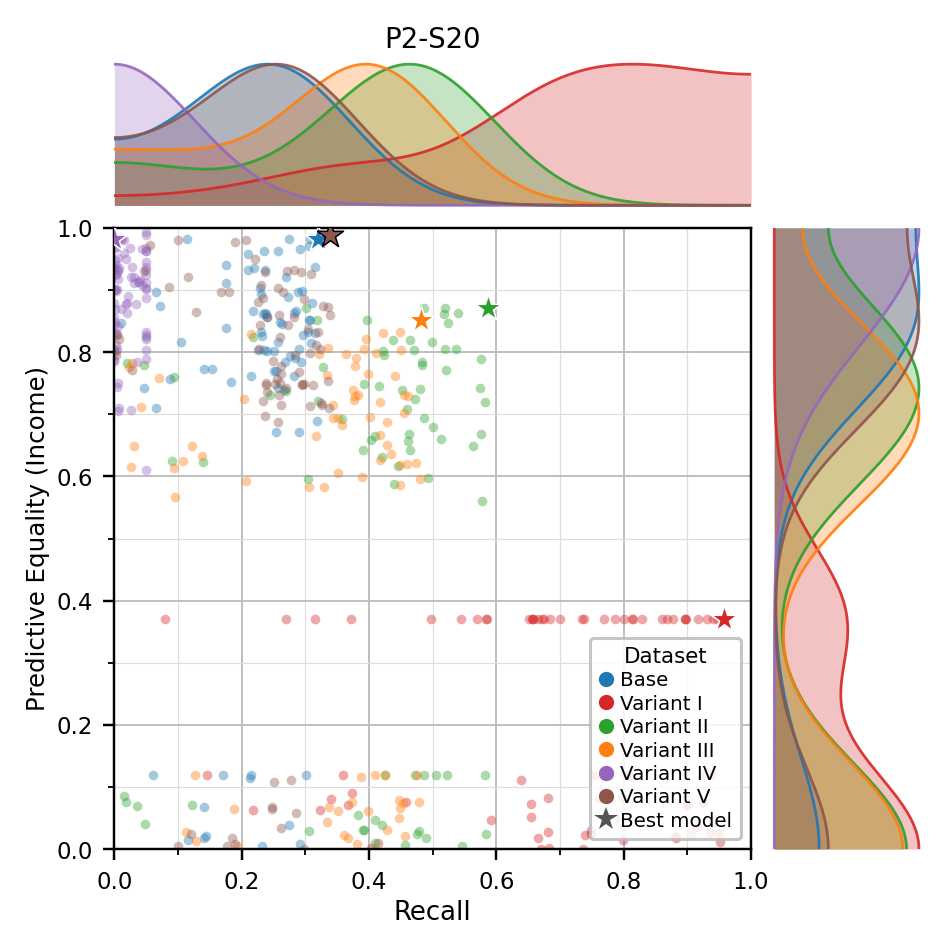}
    \caption{P2-S20 -- Income}
\end{subfigure}
\hfill
\begin{subfigure}{0.30\linewidth}
    \centering
    \includegraphics[width=\linewidth]{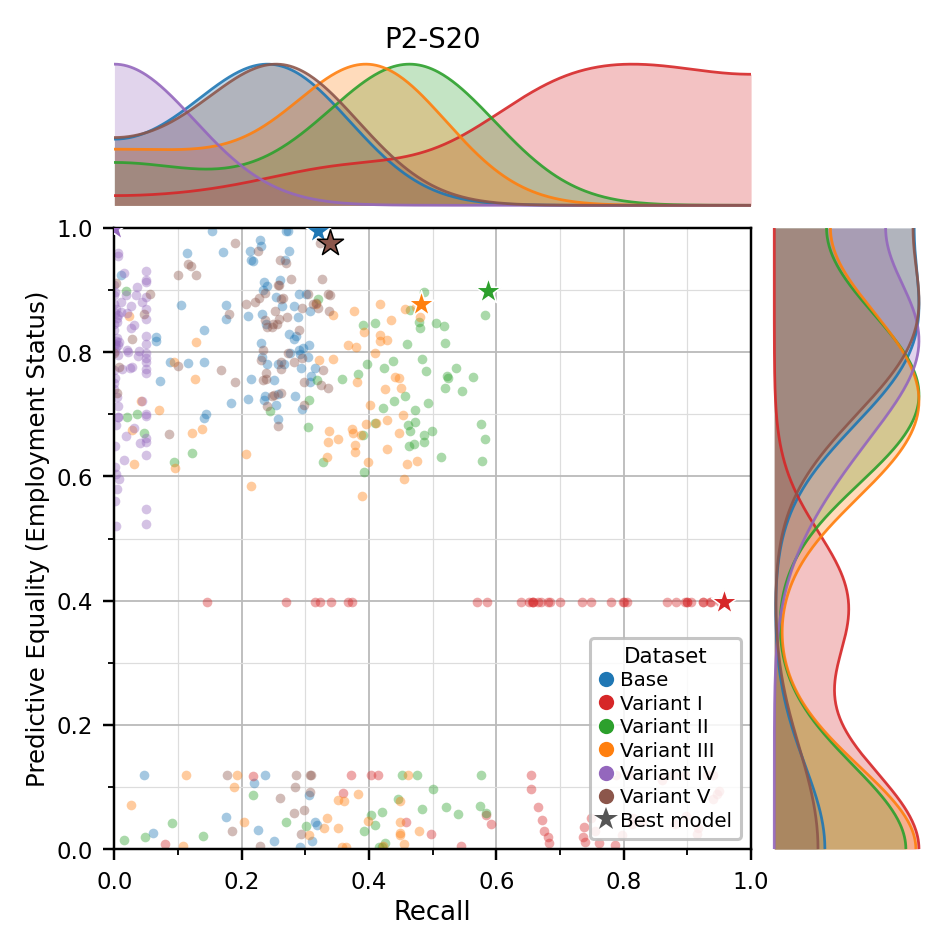}
    \caption{P2-S20 -- Employment Status}
\end{subfigure}
\\[0.5ex]
\begin{subfigure}{0.30\linewidth}
    \centering
    \includegraphics[width=\linewidth]{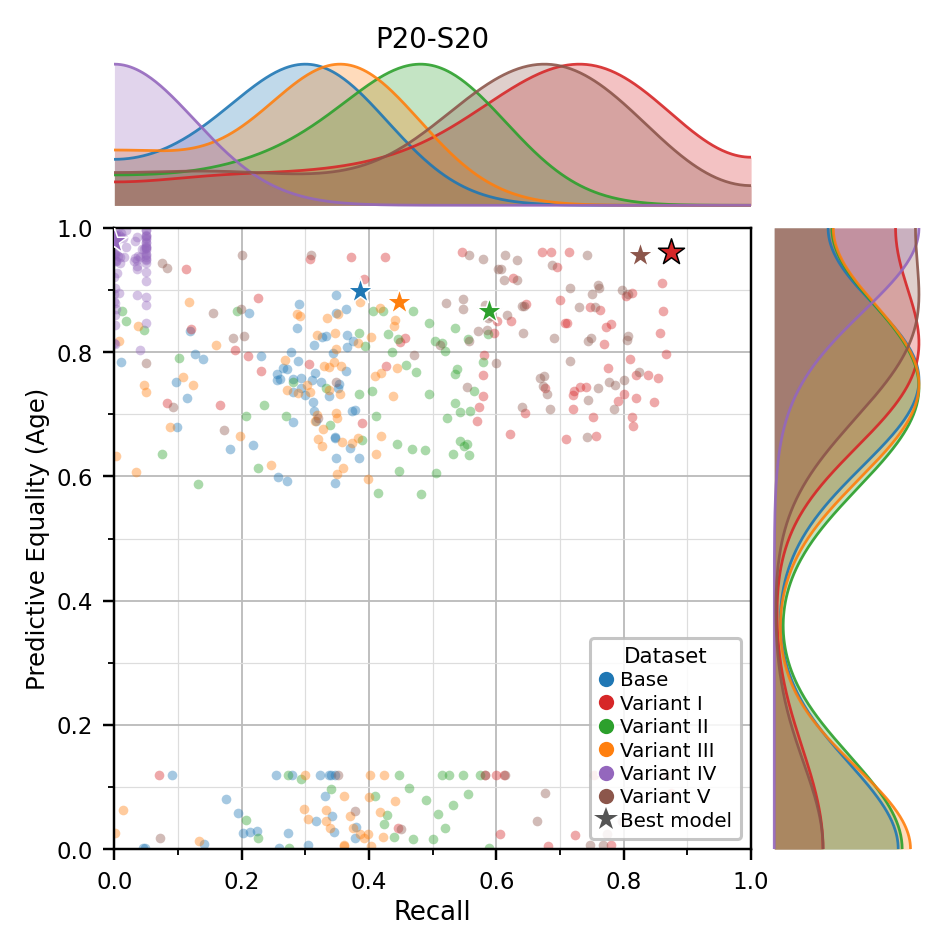}
    \caption{P20-S20 -- Age}
\end{subfigure}
\hfill
\begin{subfigure}{0.30\linewidth}
    \centering
    \includegraphics[width=\linewidth]{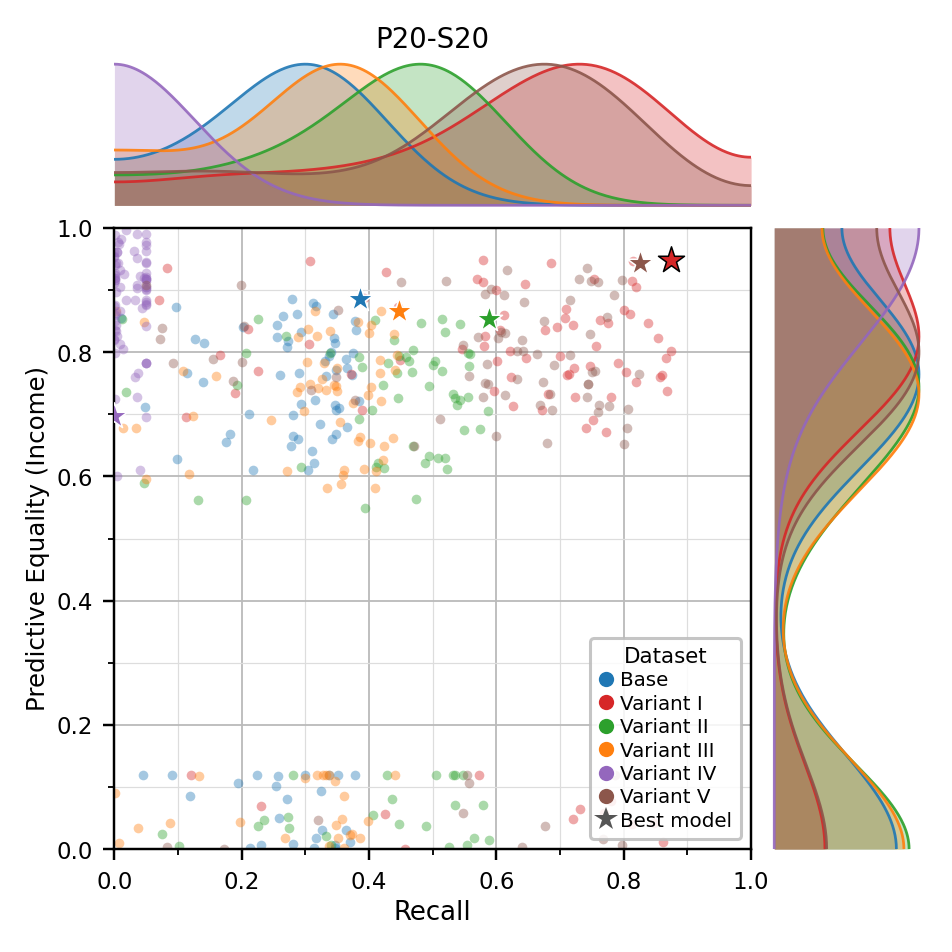}
    \caption{P20-S20 -- Income}
\end{subfigure}
\hfill
\begin{subfigure}{0.30\linewidth}
    \centering
    \includegraphics[width=\linewidth]{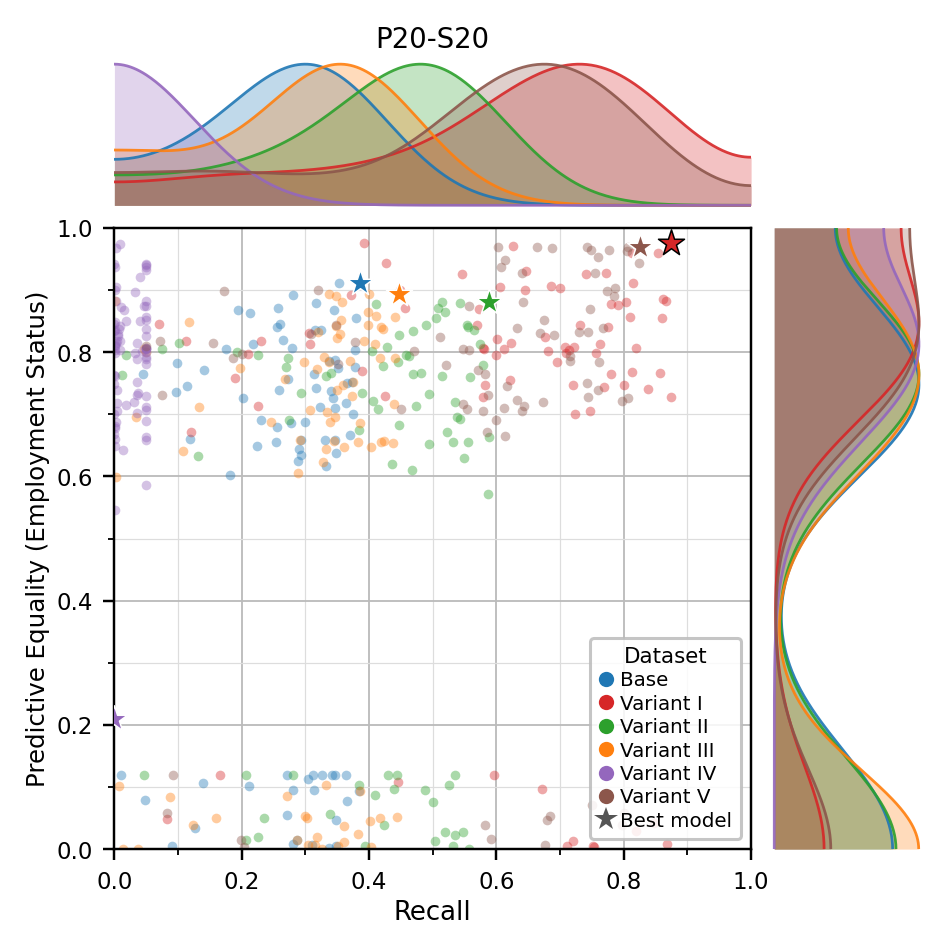}
    \caption{P20-S20 -- Employment Status}
\end{subfigure}
\\[0.5ex]
\begin{subfigure}{0.30\linewidth}
    \centering
    \includegraphics[width=\linewidth]{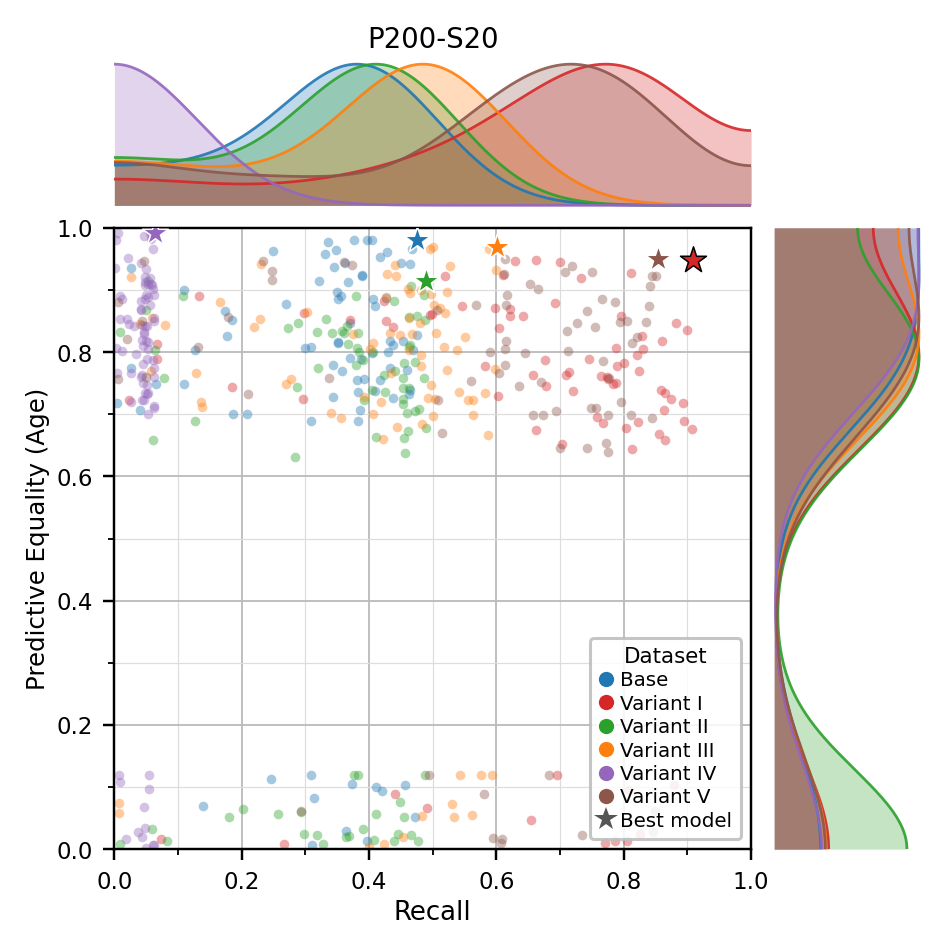}
    \caption{P200-S20 -- Age}
\end{subfigure}
\hfill
\begin{subfigure}{0.30\linewidth}
    \centering
    \includegraphics[width=\linewidth]{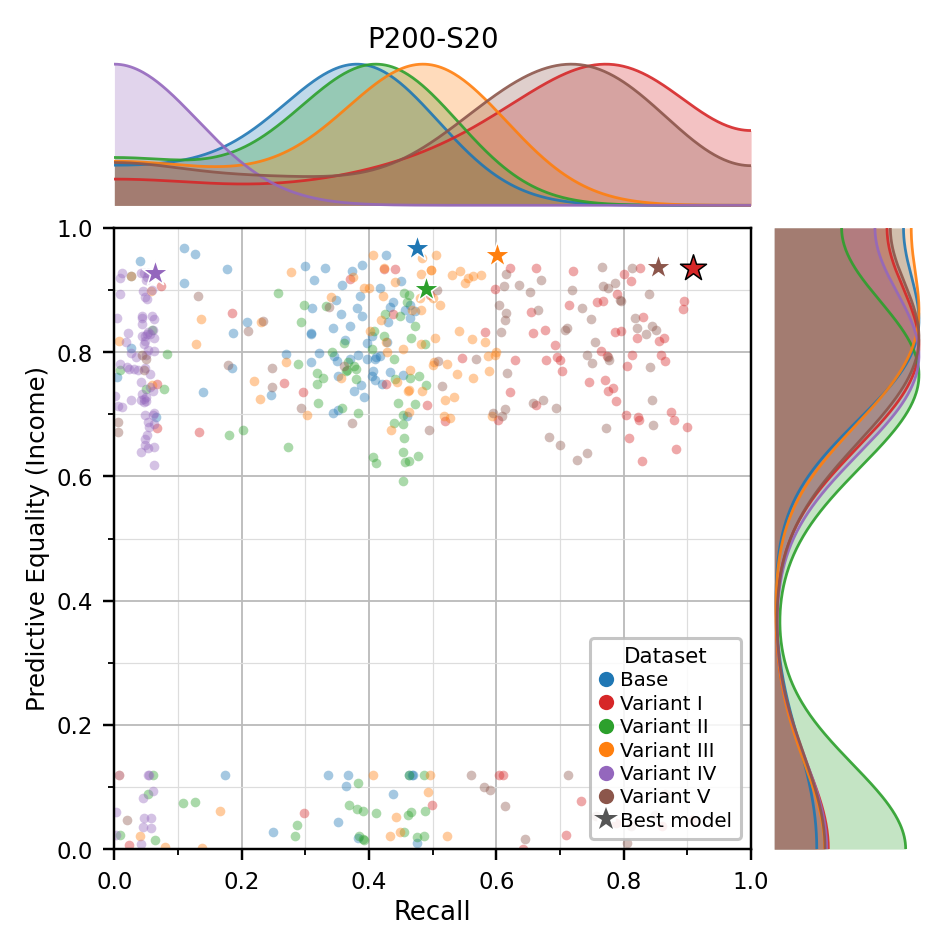}
    \caption{P200-S20 -- Income}
\end{subfigure}
\hfill
\begin{subfigure}{0.30\linewidth}
    \centering
    \includegraphics[width=\linewidth]{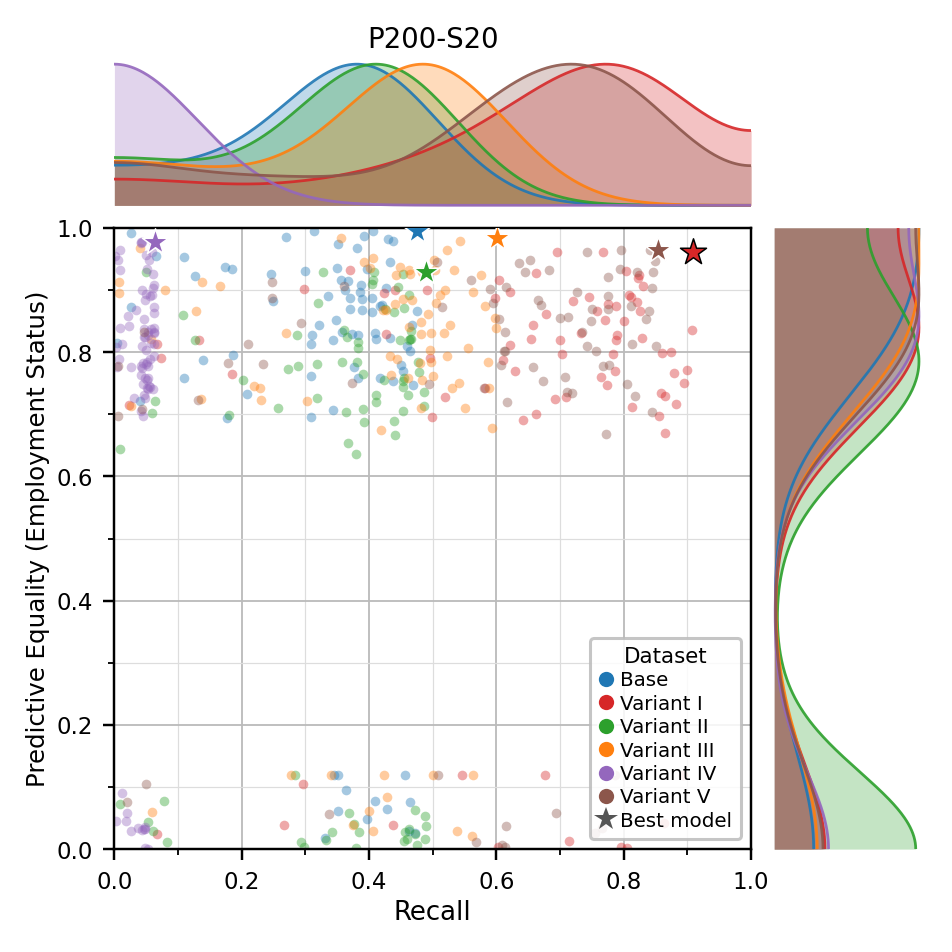}
    \caption{P200-S20 -- Employment Status}
\end{subfigure}
\caption{Evaluation of model performance (recall) and fairness (predictive equality) across varying population sizes $P \in \{2, 20, 200\}$ and sensitive attributes at time step $S = 20$.}
\label{fig:fairness}
\end{figure}

We evaluated the interplay between model performance (Recall) and fairness (Predictive Equality) by varying the neuron population size ($P \in \{2, 20, 200\}$) at a fixed temporal step ($S=20$), with results shown in Figure~\ref{fig:fairness}. The analysis covers three sensitive attributes across all six dataset variants, assessing the model's robustness to distributional shifts. The results reveal a clear and systematic improvement as population size increases. At a minimal population ($P=2$), the model exhibits significant instability. Data points are widely dispersed, with predictive equality for Age and Employment Status dropping below 0.90 for several variants. The broad marginal distributions confirm this high variance, indicating that the low-capacity model lacks the representational power to learn consistently fair and effective decision boundaries. Increasing the population to $P=20$ substantially mitigates this variance. Data points begin to cluster in the desirable top-right quadrant, showing a simultaneous improvement in both recall and fairness. This trend culminates at $P=200$, where the model achieves highly robust and stable performance. Here, the data points form a tight cluster with predictive equality values consistently at or near 1.0 across all attributes and variants. As evidenced by the sharply peaked marginal distributions, this high degree of fairness is achieved without compromising recall, which remains consistently high. In other words, the figure provides strong evidence that increasing the neuron population size is a critical factor for improving both the performance and fairness of the CSNPC architecture. A larger population endows the model with the representational capacity to learn features that are predictive yet invariant to sensitive attributes. The clear convergence from a volatile outcome at $P=2$ to a stable, high-performing, and ethically aligned result at $P=200$ validates population coding as an effective strategy for mitigating bias against the challenging distributional shifts present in the BAF dataset variants.

\FloatBarrier
\begin{figure}[pos=h]
\centering
\begin{subfigure}{0.30\linewidth} \centering
    \includegraphics[width=\linewidth]{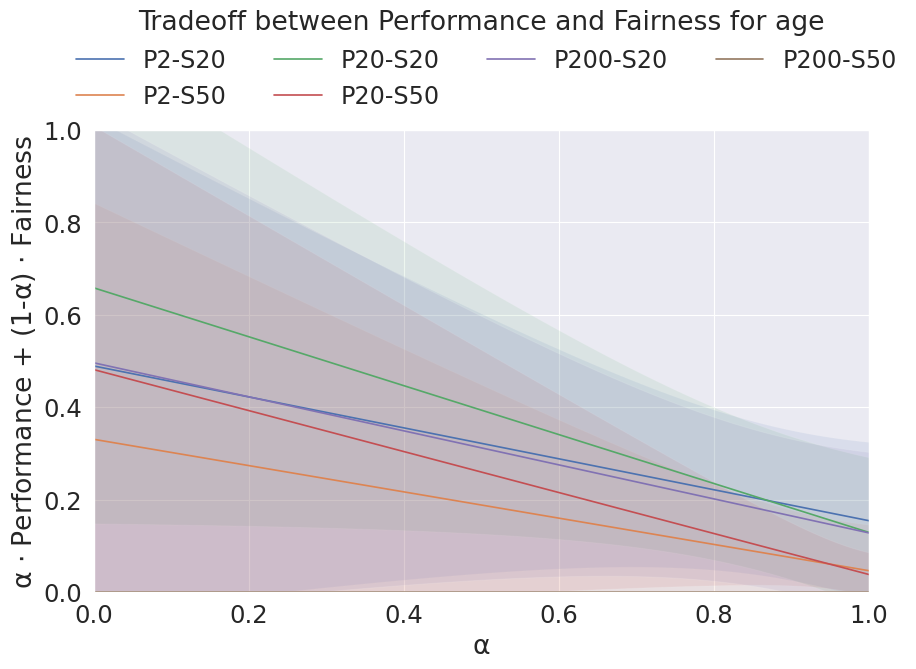}
    \caption{Age}
\end{subfigure}
\hfill
\begin{subfigure}{0.30\linewidth} \centering
    \includegraphics[width=\linewidth]{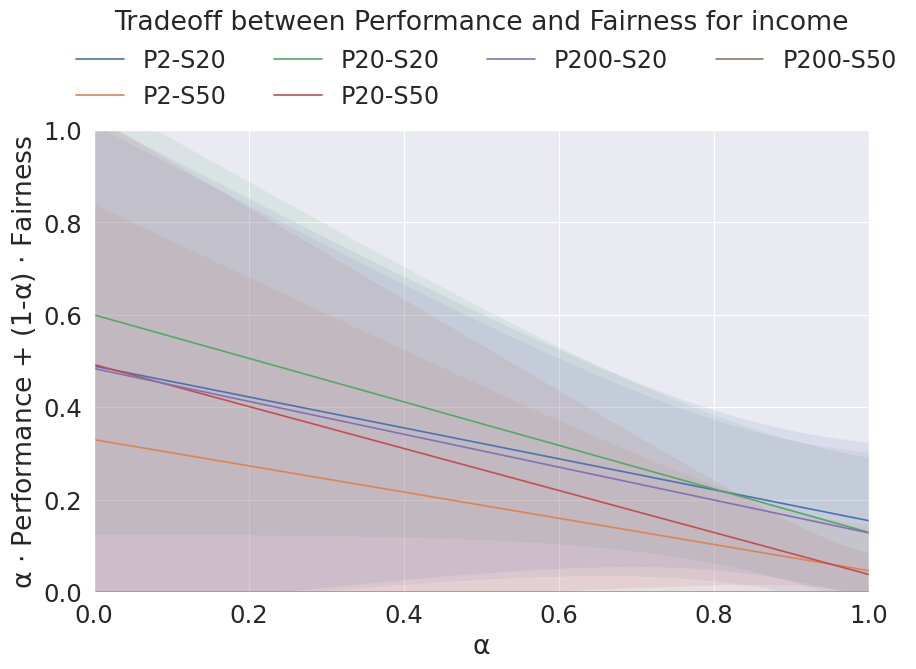}
    \caption{Income}
\end{subfigure}
\hfill
\begin{subfigure}{0.30\linewidth} \centering
    \includegraphics[width=\linewidth]{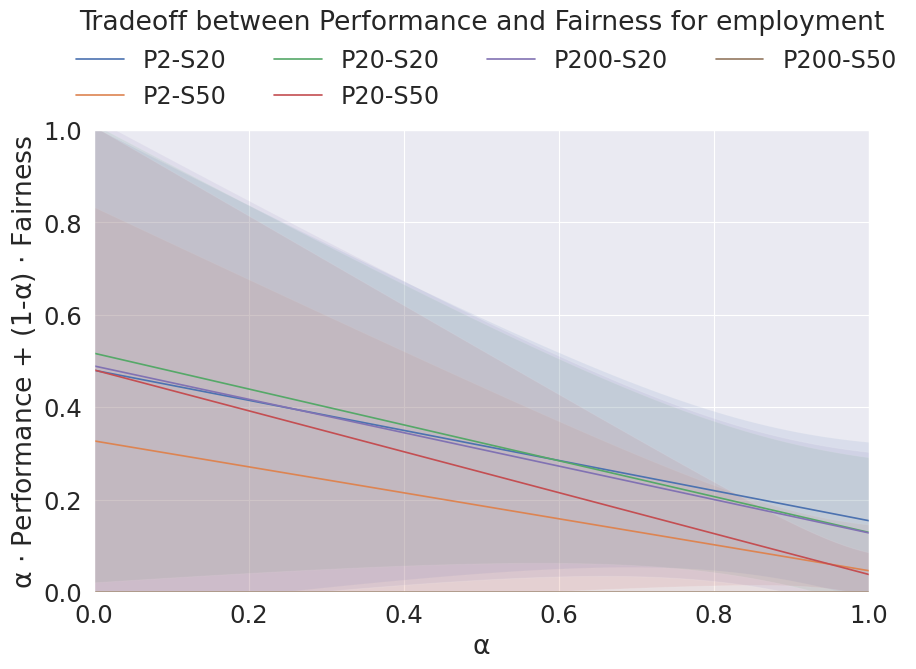}
    \caption{Employment Status}
\end{subfigure}
\caption{Fairness-performance trade-offs for sensitive attributes across spiking configurations.}
\label{fig:tradeoff}
\end{figure}

To complement this instance-level view, Figure~\ref{fig:tradeoff} aggregates fairness-performance trade-offs across all six configurations. For each attribute, increasing both population size and temporal resolution (\(P=200, S=50\)) shifts the trade-off curve favorably, yielding higher fairness at a given recall level. The trade-offs in all our approaches exhibit a negative slope, indicating a stronger emphasis on fairness over performance. Conversely, low-capacity configurations (e.g., P2-S20) display a steeper trade-off: maintaining fairness requires substantial sacrifices in recall. These findings confirm that fairness can be systematically optimized within the CSNPC architecture via RHOSS-guided configuration selection. So, the configuration P200-S20 emerges as a reliable balance point, achieving predictive equality above 0.95 across most variants while maintaining business-constrained recall levels above 0.85.

\FloatBarrier
\subsection{Comparison with Conventional Models}
Table~\ref{tab:comparison} compares CSNPC against representative spiking and non-spiking baselines under the 5\% FPR constraint. Deep SNNs trained using backpropagation through time (BPTT) and Truncated Backpropagation Through Time (TBPTT), including one- and two-layer DSNNs~\citep{dylan_bptt}, and the convolutional SpikeConv model, achieved lower recall and accuracy despite similar FPR. LightGBM~\citep{dylan_bptt}, as a strong classical baseline, reached a maximum recall of 45\%, but showed high instability in fairness metrics, with predictive equality ranging from 12\% to 77\% (age), 39\% to 52\% (income), and 11\% to 18\% (employment). The original CSNN~\citep{dylan_spiking_csnnpc} variant achieved 47.08\% recall at 4.32\% FPR with moderate fairness. In contrast, our RHOSS-optimized CSNPC attained a recall of 90.8\% and an accuracy of 98.60\%, the highest among all models, while maintaining fairness above 98\% across all sensitive attributes, clearly demonstrating better performance and ethical alignment.

\FloatBarrier
\subsection{XAI utilizing MoSSTI} 
\begin{figure}[pos=h]
\centering
\begin{subfigure}{0.30\linewidth} \centering
    \includegraphics[width=\linewidth]{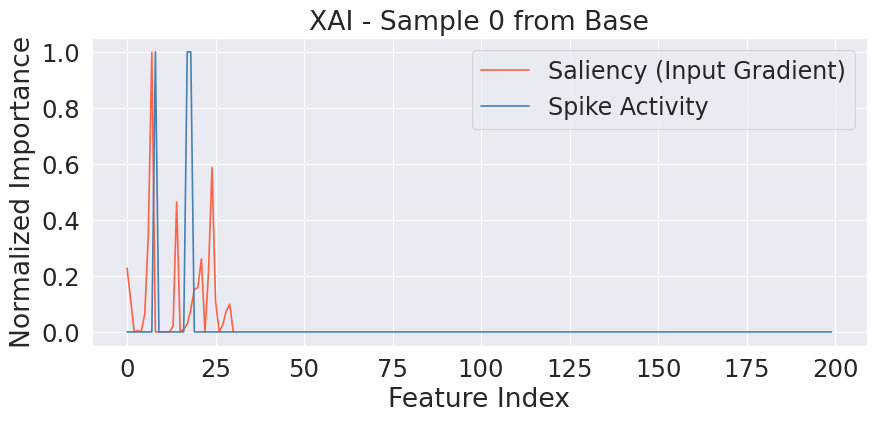}
    \caption{Sample 0 -- Base}
\end{subfigure}
\hfill
\begin{subfigure}{0.30\linewidth} \centering
    \includegraphics[width=\linewidth]{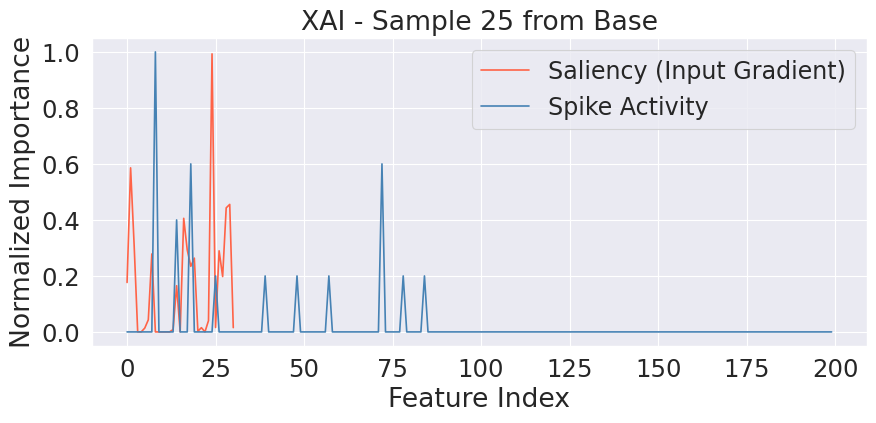}
    \caption{Sample 25 -- Base}
\end{subfigure}
\hfill
\begin{subfigure}{0.30\linewidth} \centering
    \includegraphics[width=\linewidth]{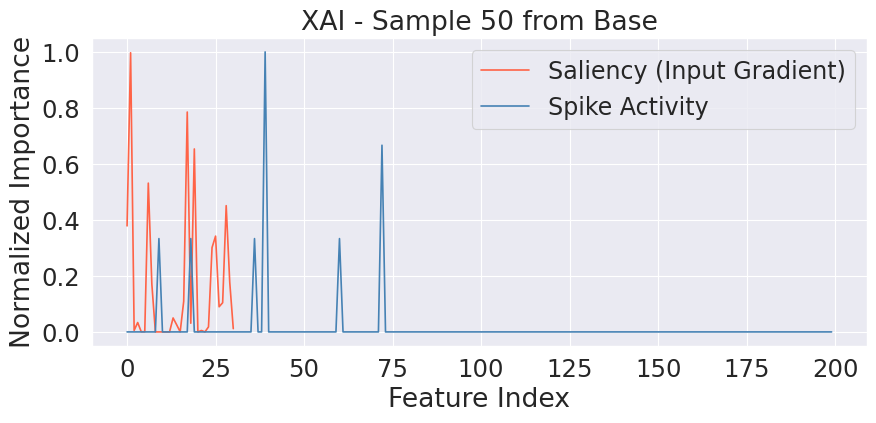}
    \caption{Sample 50 -- Base}
\end{subfigure}
\\[0.5ex]
\begin{subfigure}{0.30\linewidth} \centering
    \includegraphics[width=\linewidth]{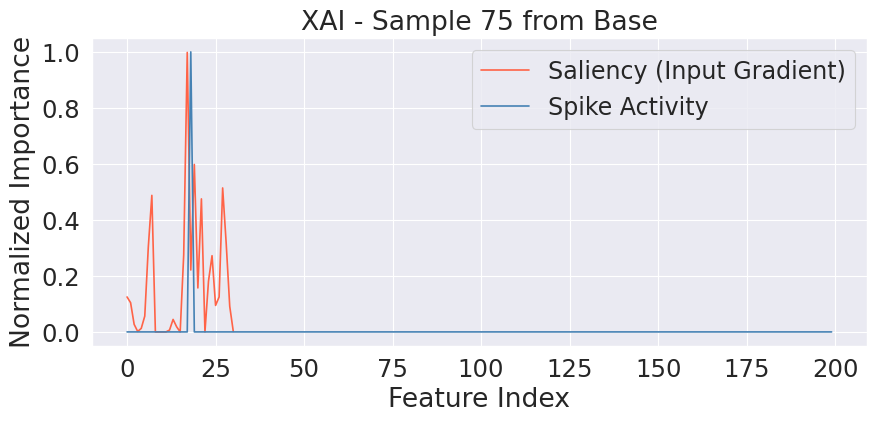}
    \caption{Sample 75 -- Base}
\end{subfigure}
\hspace{1em}
\begin{subfigure}{0.30\linewidth} \centering
    \includegraphics[width=\linewidth]{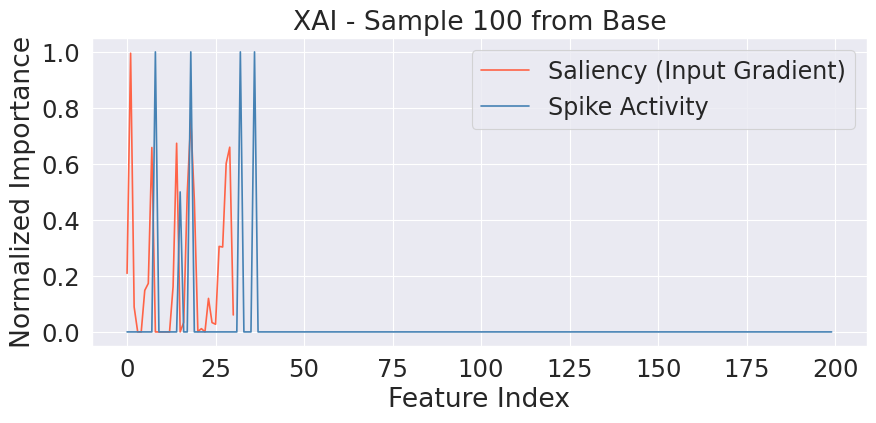}
    \caption{Sample 100 -- Base}
\end{subfigure}
\caption{Spike activity patterns for selected samples from the Base dataset.}
\label{fig:spike_activity}
\end{figure}

Our dual-path interpretability combines gradient-based saliency with spike activity profiling. This alignment serves as a proxy for faithfulness, as prior studies indicate that consistency between attribution signals correlates with user trust \cite{Hsu04052023,hoffman2019,holzinger2022}. While user studies are beyond the scope of this approach, the proposed method offers transparent reasoning grounded in both classical and biologically inspired paradigms. An explainability analysis, presented in Figure~\ref{fig:spike_activity}, was performed to interpret the model's decision-making process. 

The figure visualizes feature importance for representative input samples by juxtaposing two distinct metrics: \textbf{(i)} saliency derived from input gradients, a standard post-hoc attribution method, and \textbf{(ii)} the total spike count per feature, an intrinsic measure of neural processing in our SNN. The primary observation is the high degree of correspondence between the two importance profiles. For each sample, the features highlighted by high saliency values are precisely those that elicit sustained spike responses from the network's neuron populations. This consistency is a key finding, as it provides dual-source validation for the model's feature attribution. It confirms that the abstract importance assigned by backpropagation aligns directly with the concrete, temporal dynamics of neural firing. The observed variations in active feature sets across samples (0, 25, 50, 75, and 100) further illustrate that the learned representations are not static but are adaptively modulated by the input, reflecting the model's capacity for instance-specific reasoning. \begin{figure}[pos=h]
\centering
\begin{subfigure}{0.30\linewidth} \centering
    \includegraphics[width=\linewidth]{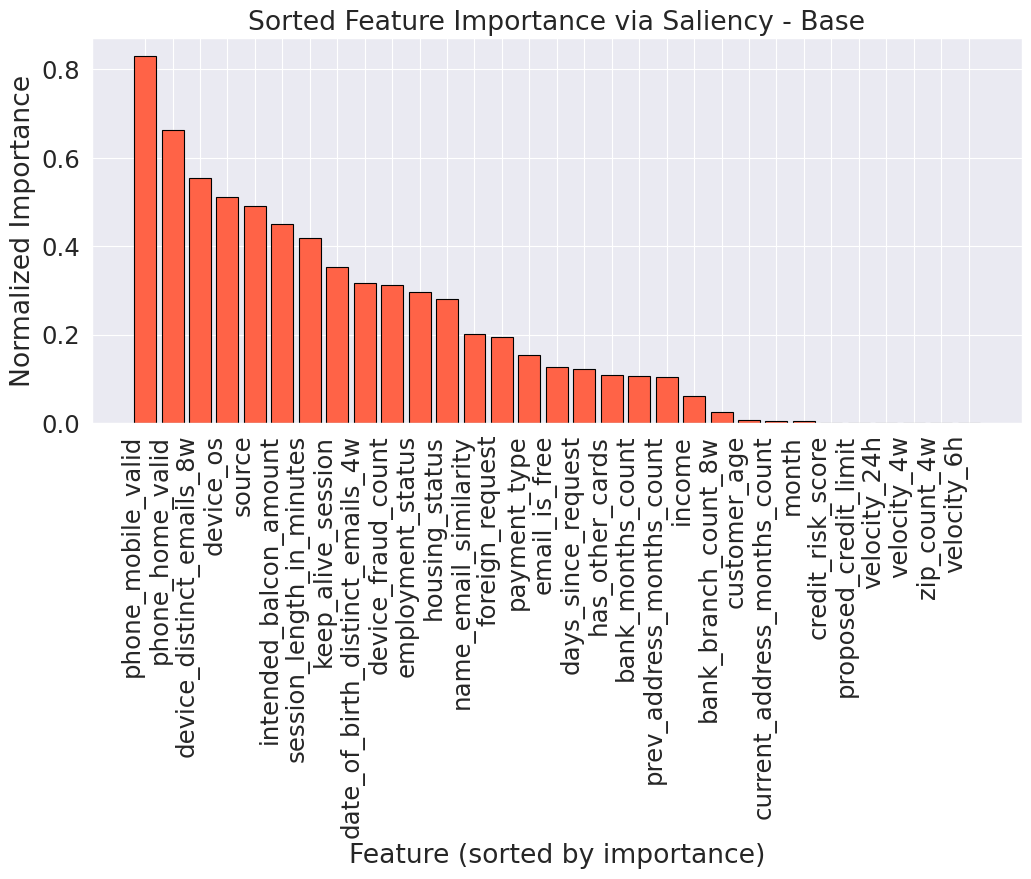}
    \caption{Base}
\end{subfigure}
\hfill
\begin{subfigure}{0.30\linewidth} \centering
    \includegraphics[width=\linewidth]{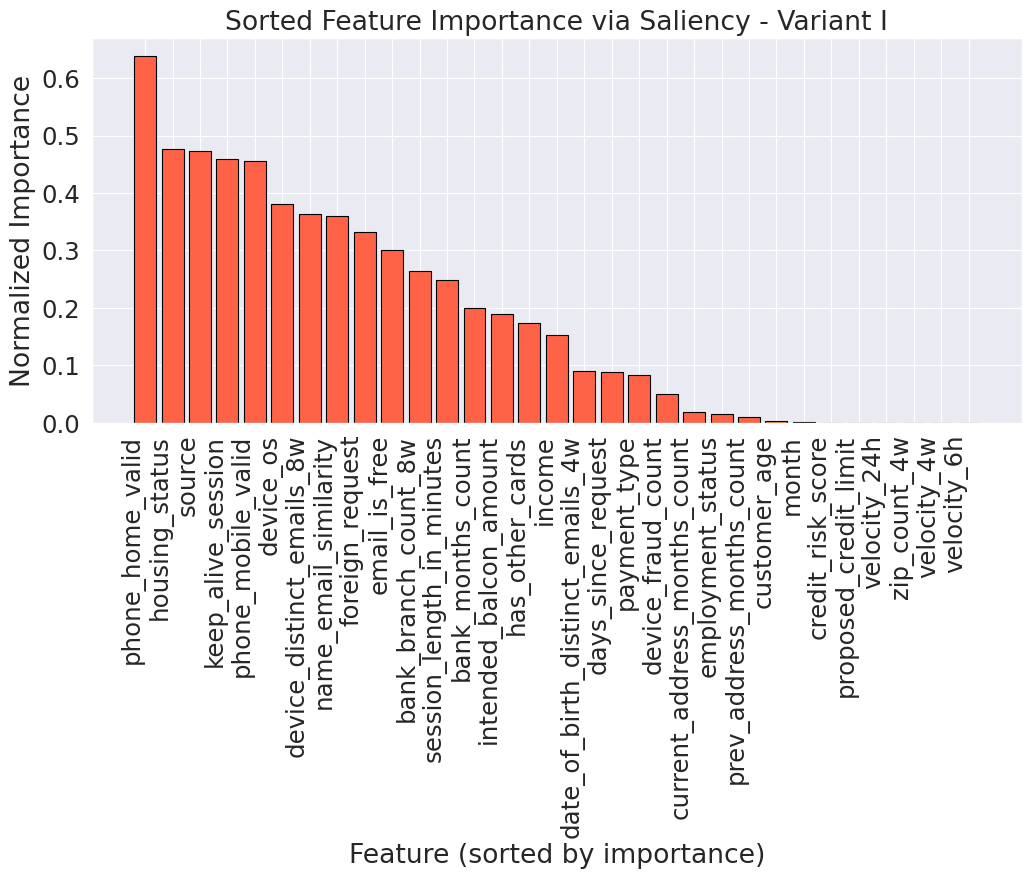}
    \caption{Variant I}
\end{subfigure}
\hfill
\begin{subfigure}{0.30\linewidth} \centering
    \includegraphics[width=\linewidth]{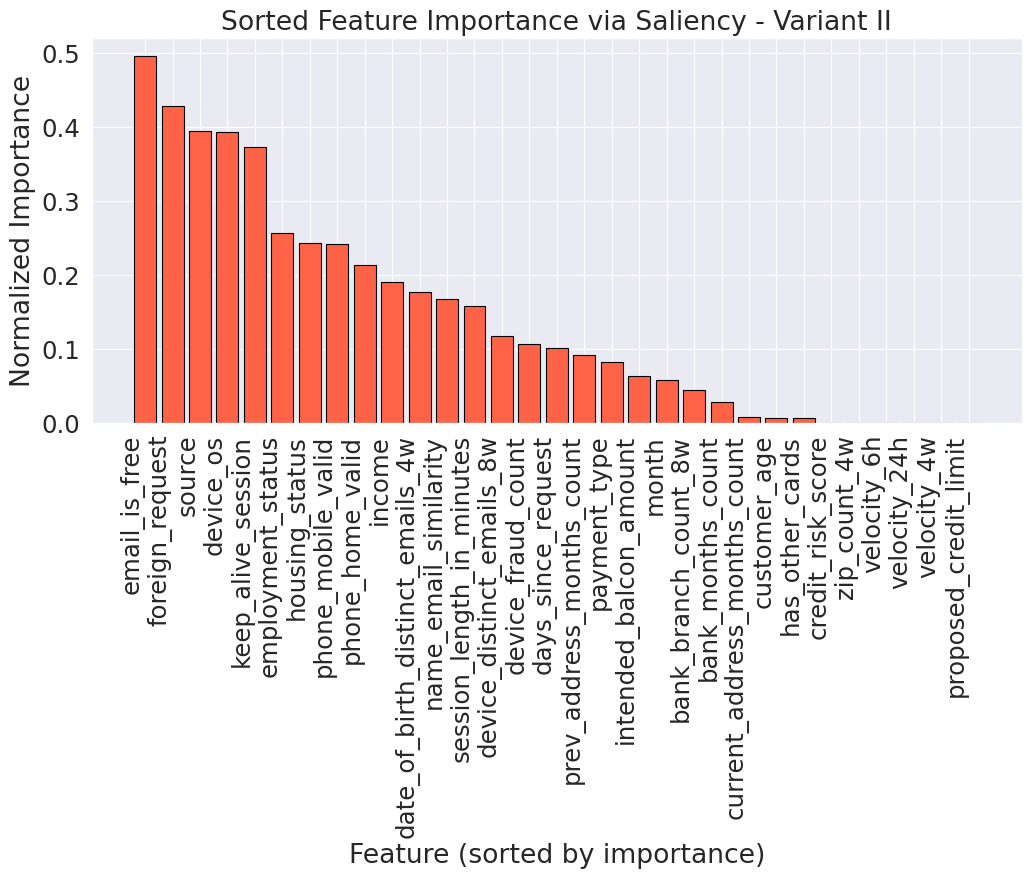}
    \caption{Variant II}
\end{subfigure}
\\[0.5ex]
\begin{subfigure}{0.30\linewidth} \centering
    \includegraphics[width=\linewidth]{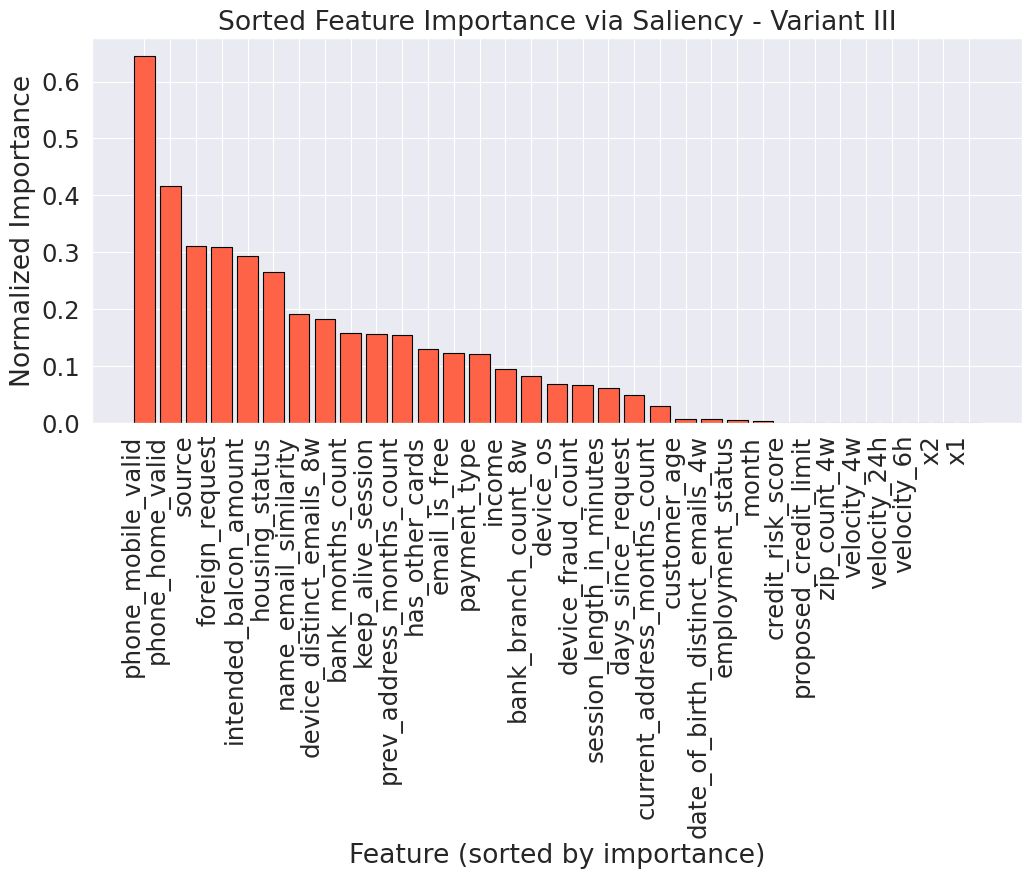}
    \caption{Variant III}
\end{subfigure}
\hfill
\begin{subfigure}{0.30\linewidth} \centering
    \includegraphics[width=\linewidth]{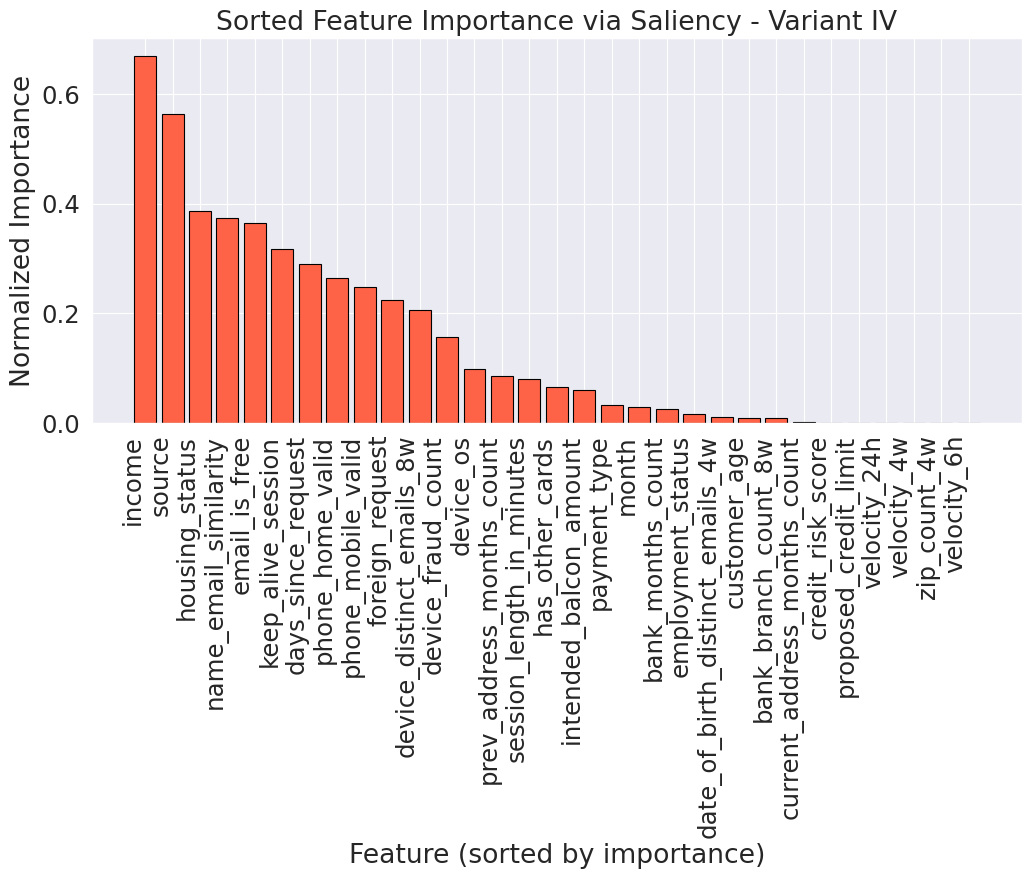}
    \caption{Variant IV}
\end{subfigure}
\hfill
\begin{subfigure}{0.30\linewidth} \centering
    \includegraphics[width=\linewidth]{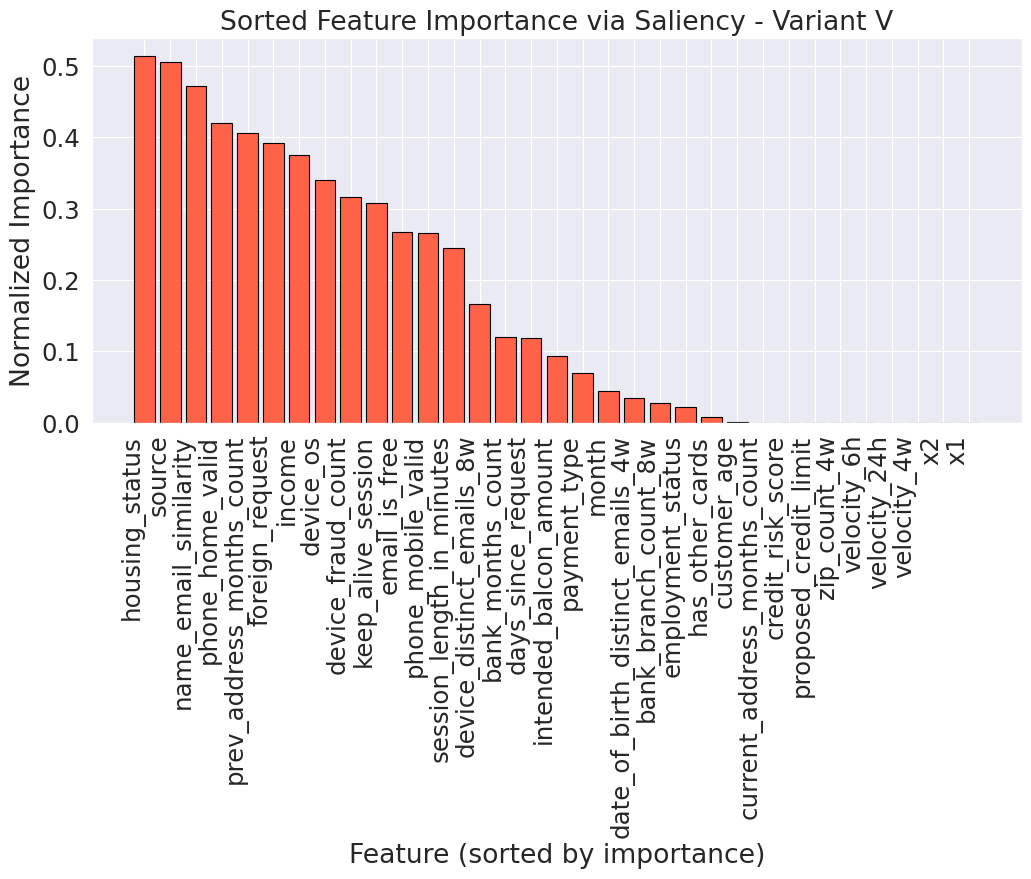}
    \caption{Variant V}
\end{subfigure}
\caption{Feature importance visualized via saliency maps across dataset variants.}
\label{fig:feature_importance}
\end{figure}

Figure~\ref{fig:feature_importance} further depicts feature importance obtained via saliency maps across dataset variants. The feature importance distributions vary slightly between the variants, with the Base dataset showing a higher weight for "phone\_home\_valid" and "device\_phone\_valid," while in other variants, the importance shifts slightly toward "email\_is\_free" and "housing\_status." Despite these differences, the model consistently identifies transaction volume, account tenure, and income level as dominant indicators of fraud. Saliency explanations remain stable across biased variants, reinforcing CSNPC's capacity for resilient, explainable, and ethically aware learning.

\subsection{Computational Cost Analysis and Trade-offs}

\begin{table}[!t]
\centering
\caption{Computational Cost Breakdown (P200-S20 Configuration)}
\label{tab:cost_breakdown}
\begin{tabular}{lccc}
\hline
\textbf{Phase} & \textbf{Time (minutes)} & \textbf{Energy (kWh)*} & \textbf{Operations} \\
\hline
RHOSS Optimization & $\approx$598 & 4.2 & Q-table updates, heuristic selection \\
SNN Training (per trial) & $\approx$6.8 & 0.048 & Forward passes, gradient computation \\
Final Model Training & $\approx$112 & 0.79 & Full 20-epoch training \\
\textbf{Total Training} & $\approx$\textbf{711} & \textbf{5.0} & \textbf{\(\sim10^9\) operations} \\
Inference (100K samples) & $\approx$13.0 & 0.0065 & Forward pass only \\
\textbf{Per-sample Inference} & $\approx$\textbf{0.0078 sec} & \textbf{\(6.5\times10^{-8}\) kWh} & \textbf{\(\sim10^3\) operations} \\
\hline
\end{tabular}

\smallskip
\footnotesize{*Estimated for NVIDIA T4 GPU (70W TDP).}
\end{table}

\begin{table}[!t]
\centering
\caption{Training Time vs. Performance Comparison}
\label{tab:time_vs_perf}
\begin{tabular}{lccc}
\hline
\textbf{Model} & \textbf{Training Time (min)} & \textbf{Recall@5\%FPR} & \textbf{Time per Recall Point} \\
\hline
CSNPC+RHOSS (Ours) & $\approx$711 & 0.908 & 783 min \\
SpikeConv M~\citep{dylan_bptt} & $\approx$680 & 0.570 & $\approx$1193 min \\
CSNN~\cite{dylan_spiking_csnnpc} & $\approx$1425 & 0.471 & $\approx$3025 min \\
LightGBM~\citep{dylan_bptt} & $\approx$15 & 0.450 & $\approx$33 min \\
\hline
\end{tabular}
\end{table}

\subsubsection{Cost-Benefit Analysis}
As shown in Table~\ref{tab:cost_breakdown}, the total training time is approximately 12~hours, incurred only once during model development. This cost becomes negligible when amortized over millions of deployment transactions. During inference, the model achieves a latency of 7.8~ms per sample, enabling throughput of about 128~samples/s, comparable to LightGBM’s 145~samples/s but with nearly twice the recall (Table~\ref{tab:time_vs_perf}). The estimated training energy consumption of 5~kWh is also efficiently amortized. With an inference cost of \(6.5\times10^{-8}\)~kWh per sample, processing one million transactions requires only 0.065~kWh, significantly lower than the \(\sim0.15\)~kWh typically required by equivalent ANN models. Hence, the overall computational cost remains marginal in large-scale deployments.

\subsubsection{When Cost is Justified}
These computational costs are particularly warranted in high-value applications such as banking and insurance fraud detection, where a 40\% improvement in recall can prevent millions of dollars in losses. RHOSS additionally ensures fairness-aware optimization, a key requirement for regulatory compliance. Moreover, the model’s stable performance allows for quarterly retraining, thereby minimizing long-term costs.

\subsubsection{Comparison with Simpler Methods}
Although LightGBM trains roughly 47$\times$ faster, its recall is 50\% lower. As summarized in Table~\ref{tab:time_vs_perf}, this performance difference translates into substantial financial implications. In a banking scenario processing 10 million transactions daily with an average fraud loss of \$500, our model (recall 90.8\%) would prevent 
\$4.54~million in losses per day, compared to \$2.25~million with LightGBM (recall 45\%). The additional 12 hours of training thus yields an extra \$2.29~million in daily savings, achieving a return on investment within less than one hour of deployment. Therefore, the computational overhead of CSNPC+RHOSS is empirically and economically justified for high-stakes, fairness-sensitive applications.

\section{Conclusions}
\label{sec:conclude}
In this study, we proposed the application of SNNs with hyper-heuristic optimizers and XAI, focusing on handling highly imbalanced datasets. Our architecture design and hyperparameter optimization enabled a novel approach that effectively addresses fraud detection while adhering to strict business constraints on FPRs. A comparative analysis with previous SNNs and older models, such as decision trees and gradient boosting algorithms, showed that our model achieved competitive performance, outperforming even the latest SNN models by a margin. Furthermore, our CSNPC-based model also exhibited fairness by maintaining predictive equality across sensitive customer attributes, such as age, income, and employment status, which are critical in preventing discriminatory practices. Finally, our approach with XAI ensured that the model remains understandable to its clients, ensuring fairness and transparency. 

Still, the model is not without its flaws and has some areas for improvement. While our explainability module provides visual comparisons between saliency maps and spike activity, it does not include quantitative faithfulness metrics or user studies to confirm that non-experts can reliably interpret the model’s decisions. Additionally, the current evaluation relies on the synthetic BAF dataset, which, although designed to emulate real-world bias structures, may not fully capture the complexity of live banking environments. The offline hyperparameter search using RHOSS introduces significant computational overhead, which, while amortized over deployment, remains a practical limitation for rapid prototyping. Finally, fairness is assessed primarily through predictive equality; other fairness notions, such as equal opportunity or demographic parity, were not explored. Future work will address these limitations by: (i) incorporating quantitative explainability metrics such as deletion/insertion AUC and comprehensiveness scores; (ii) conducting user studies to validate interpretability for non-experts; (iii) extending evaluation to proprietary banking datasets under strict privacy constraints; (iv) optimizing RHOSS for reduced search cost through parallelization or transfer learning; and (v) broadening fairness objectives to include multiple definitions and intersectional group analysis.

\section*{Data and Code Availability}
The \textit{Bank Account Fraud Dataset Suite (NeurIPS 2022)} is publicly available on Kaggle (\url{https://www.kaggle.com/datasets/sgpjesus/bank-account-fraud-dataset-neurips-2022}) under the \textit{CC BY-NC-ND 4.0} license for academic and non-commercial use. We used all six variants (Base, Variant I-V) with the standard 90-10 train-test split as distributed.

\section*{Competing Interest Statement}
The authors declare that they have no known competing financial interests or personal relationships that could have appeared to influence the work reported in this paper.

\section*{Funding Sources}
This research did not receive any specific grant from funding agencies in the public, commercial, or not-for-profit sectors.

\printcredits

\bibliographystyle{elsarticle-num}
\bibliography{main}

\end{document}